%% file: acl_latex.tex
\pdfoutput=1

\documentclass[11pt]{article}
\usepackage{float}

\usepackage{acl}

\usepackage[utf8]{inputenc}
\usepackage{pgfplots}
\usepackage{dsfont}

\DeclareUnicodeCharacter{2212}{−}
\usepgfplotslibrary{groupplots,dateplot}
\usetikzlibrary{patterns,shapes.arrows}
\pgfplotsset{compat=newest}

\usepackage{tikzscale}
\usepackage{amsmath}
\newcommand{\probP}{\text{I\kern-0.15em P}}

\usepackage{multirow, colortbl}
\usepackage{color}
\usepackage{array}
\usepackage{multirow}
\usepackage{pifont}
\usepackage{CJKutf8}
\usepackage{tabularx,booktabs}
\usepgfplotslibrary{groupplots}
\usepackage{makecell}
\usepackage{enumitem}
\usepackage{placeins}

\usepackage[normalem]{ulem}
\useunder{\uline}{\ul}{}

\usepackage{graphicx}
\usepackage{booktabs}

\usepackage{pifont}

\usepackage{pgfplots}
\pgfplotsset{width=10cm,compat=1.9}

\definecolor{ablation6}{HTML}{fcefed}
\definecolor{ablation_tie}{HTML}{fce3e1}

\definecolor{ablation5}{HTML}{fcd8d4}
\definecolor{ablation4}{HTML}{FBC3BC}
\definecolor{ablation3}{HTML}{F7A399}
\definecolor{ablation2}{HTML}{F38375}
\definecolor{ablation1}{HTML}{EF6351}

\usepackage[]{algpseudocode}
\usepackage[]{algorithm}
\usepackage{float}
\algtext*{EndFor}%
\algtext*{EndProcedure}%

\usepackage{booktabs}
\usepackage[normalem]{ulem}
\useunder{\uline}{\ul}{}

\usepackage{times}
\usepackage{latexsym}
\usepackage{adjustbox}

\usepackage[T1]{fontenc}
\usepackage[utf8]{inputenc}
\usepackage[russian,english]{babel}
\newcommand{\russian}[1]{{\fontencoding{T2A}\selectfont\foreignlanguage{russian}{#1}}}

\usepackage{amsmath}
\usepackage{amssymb}
\usepackage{booktabs}
\usepackage{multirow}
\usepackage{scalerel,xparse}
\usepackage[utf8]{inputenc}
\usepackage{cleveref}
\usepackage{microtype}
\usepackage{enumitem}
\crefformat{section}{\S#2#1#3}
\crefformat{subsection}{\S#2#1#3}
\crefformat{subsubsection}{\S#2#1#3}

\input{xling-write}

\title{Guiding Large Language Models to Post-Edit \\ Machine Translation with Error Annotations}

\author{Dayeon Ki \\
  Computer Science \\
  University of Maryland \\
  \texttt{dayeonki@cs.umd.edu} \\\And
  Marine Carpuat \\
  Computer Science, UMIACS \\
  University of Maryland \\
  \texttt{marine@cs.umd.edu} \\}

\begin{document}
\maketitle

\begin{abstract}
Machine Translation (MT) remains one of the last NLP tasks where large language models (LLMs) have not yet replaced dedicated supervised systems.
This work exploits the complementary strengths of LLMs and supervised MT by guiding LLMs to automatically post-edit MT with external feedback on its quality, derived from Multidimensional Quality Metric (MQM) annotations. Working with LLaMA-2 models, we consider prompting strategies varying the nature of feedback provided and then fine-tune the LLM to improve its ability to exploit the provided guidance. Through experiments on Chinese-English, English-German, and English-Russian MQM data, we demonstrate that prompting LLMs to post-edit MT improves TER, BLEU and COMET scores, although the benefits of fine-grained feedback are not clear. Fine-tuning helps integrate fine-grained feedback more effectively and further improves translation quality based on both automatic and human evaluation.\footnote{We release our code, dataset, model checkpoints at \url{https://github.com/dayeonki/mt_feedback}.}

\end{abstract}



\input{pages/introduction}

\input{pages/related_work}

\input{pages/method}

\input{pages/experimental_setup}
\input{pages/prompting_results}

\input{pages/ft_results}

\input{pages/analysis}

\input{pages/analysis_noerror}

\input{pages/conclusion}

\input{pages/limitation}

\input{pages/acknowledgement}


\bibliography{custom,naacl2024}

\appendix

\section{Fine-grained Feedback format}
\label{sec:fine_grained}
In this section, we discuss the details on the format of fine-grained feedback both human-annotated and automatically annotated by InstructScore or xCOMET. We refer to fine-grained feedback in three components: error span position, error type, and error severity level. MQM annotations and InstructScore use the same MQM hierarchy to define error type as shown in Table \ref{tab:mqm_type}, with InstructScore omitting categories such as ``Source error'', ``Non-translation'', and ``Other''. Unlike these, xCOMET does not provide error type information in their annotation.

The levels of error severity are summarized in Table \ref{tab:severity_level}. In our prompting experiments, we eliminate instances annotated as ``No-error'' in the MQM dataset, as our focus is on understanding the role of external feedback in post-editing \textbf{erroneous} translations. However, for fine-tuning, we include all instances, regardless of their error severity level.

\input{tables/severity_level}
\input{tables/mqm_scores}

\section{Error Annotation Examples}
\label{sec:error_annotate_examples}
In Table \ref{tab:error_annotate}, we present error annotation examples from three sources: MQM, xCOMET, and InstructScore. We obtain automatic annotations of the same evaluation dataset using InstructScore and xCOMET. The consistency of these error annotations across different tools is further discussed in Section \ref{sec:dataset}.

\section{Dataset Details}
\subsection{MQM Dataset}
\label{ref:mqm_dataset_details}
We analyze 1,000 MQM data instances used for evaluation. We note that the average number of error spans per sentence is 1 as from the original MQM dataset. The average error span length is 13.5 for Zh-En, 11.3 for En-De, and 9.3 for En-Ru. Further, we observe the error type and severity level distribution in Figure \ref{fig:mqm_type_distribution} and \ref{fig:mqm_severity_distribution}. Across all language pairs, ``\textit{Accuracy}'' errors are the majority (524 for Zh-En, 362 for En-De, 592 for En-Ru), followed by ``\textit{Fluency}'' (274 for Zh-En, 324 for En-De, 257 for En-Ru). For the severity level, Zh-En has the most ``\textit{major}'' errors (512/1000), then En-Ru (388/1000) and En-De (202/1000).

\input{figures/mqm_type_distribution}

\section{Fine-tuning Details}
\subsection{Fine-tuning Dataset format}
We illustrate the instruction template used for constructing fine-tuning dataset in Table \ref{tab:finetune_data}. We explicitly include the fine-grained errors in the instruction to guide LLMs on how to leverage them as hints and align in their improved translation outputs. We employ the same instruction format during inference time.

\subsection{LLaMA-2 13B Results}
\label{sec:13_fine_tune}
We also extend the fine-tuning of LLaMA-2 13B for two settings, mirroring the 7B setup: bilingual and multilingual. We follow the identical experimental setup as in the 7B experiment. We show that similar trend is observed; fine-tuning with error annotations show better performance than the original baseline and the zero- and 10-shot prompting results across all metrics.

\input{tables/fine_tuning_13}

\section{Detailed Results}
\label{sec:additional_experiment}
\subsection{LLaMA-2 Original}
\label{sec:llama-2-original}
We show the detailed numerical results for the LLaMA-2 7B and 13B experiments in Table \ref{tab:llama-7} and \ref{tab:llama-13} respectively. In zero-shot prompting, the 13B model outperforms 7B, which can be attributable by larger LLMs having stronger instruction-following capabilities than their smaller counterparts \cite{wei2022emergent}. Further, 13B shows similar trend observed for 7B, where the use of external feedback in 10-shot prompting significantly improves performance.

\subsection{LLaMA-2 Chat}
\label{sec:llama-2-chat}
We expand our experiments with LLaMA-2-chat, an instruction fine-tuned version of LLaMA-2. Although they are optimized to better follow the instructions that users specify, we show that LLaMA-2 models consistently outperform the chat counterparts in Table \ref{tab:llama-chat}. Our findings indicate that instruction-following ability of LLMs might not be the only determining factor for successful MT post-editing.

\subsection{Fine-grained Components}
In Table \ref{tab:fine_grained}, we observe the impact of each component of the fine-grained feedback: error span position, error type, and severity level with 200 randomly sampled test cases. We examine that while the individual contribution of each error component is trivial, interestingly, providing only the severity level information consistently yields similar or superior results compared to providing all three components simultaneously. This shows that there could be other forms of feedback effective when prompted to LLMs, which we leave for future work.

\subsection{Translate from Scratch}
\label{sec:translate_from_scratch}
We present zero-shot LLaMA-2 translation results in Table \ref{tab:translate_from_scratch}. We report the scores for 1,000 WMT test instances used in our main evaluation, as translated with LLaMA-2 7B and 13B models with the prompt template as: \textit{``Translate from \{source language\} to \{target language\} without any explanation.\textbackslash n\{source language\}: \{source sentence\}\textbackslash n\{target language\}:''}. Results show that LLaMA-2 7B is not powerful at translating compared to the baseline hypothesis translations provided by the MQM. LLaMA-2 13B shows comparable results to the original performance except for En-De where it slightly outperforms the original. We would expect much higher scores if the test set had been memorized as part of the LLaMA-2 pre-training data. Further, we notice that translating from scratch with LLaMA-2 7B consistently shows lower performance than post-editing regardless of feedback types. For 13B, again translating shows lower performance compared to post-editing with generic or xCOMET feedback but similar to score-based or MQM feedback.

\subsection{Qualitative Analysis}
\label{sec:fine_tune_qualitative}
In this section, we demonstrate how fine-tuning enhances the alignment of LLM behavior with the external feedback. Tables \ref{tab:qualitative_zhen}, \ref{tab:qualitative_ende}, and \ref{tab:qualitative_enru} illustrate output translations generated by LLaMA-2 7B incorporating different types of feedback. While relatively coarse feedback (generic and score-based) are not able to accurately pinpoint and correct the targeted error spans, fine-grained feedback (MQM, InstructScore, and xCOMET) resolves this issue. Further, even in instances where fine-grained feedback falls short, fine-tuning enables the model to generate translations that more effectively narrow the gap. We also demonstrate that translations from the fine-tuned model not only resolves the errors but also makes it more natural (less translationese) in the target language.

\section{Human Evaluation}
\label{sec:human_eval_details}

\subsection{Evaluation Details}
For human evaluation, we employ Qualtrics\footnote{\url{https://www.qualtrics.com/}} to design our survey and Prolific\footnote{\url{https://www.prolific.com/}} to recruit human annotators. We randomly sample 50 instances for each language pair and further divide them into two separate sessions. Each session consists of 25 examples and is estimated to take approximately 30 minutes to complete. For every session and language pair, we engage 3 annotators who are fluent in both source and target languages. For example, when evaluating the Chinese-English pair, we choose annotators fluent in both Chinese and English. Consequently, for each language pair, we recruit a total of 6 annotators, amounting to 18 annotators overall. We offer a compensation of \$7 per session, totaling \$126 for the entire human evaluation process.

\subsection{Annotator Instructions}
In Figure \ref{fig:human_intro} and \ref{fig:human_example}, we present the instructions and survey content provided to our annotators. Each annotator reviews 25 set of examples, each consisting of the source text, Translation 1 (the initial translation), and Translation 2 (the output translation from our fine-tuned model). They are tasked with comparing these two translation on a Likert scale ranging from 0 (Strongly disagree) to 5 (Strongly agree). This comparison is based on two criteria: (1) ``\textit{Translation 2 is better than Translation 1}'' evaluates whether the output translation more effectively conveys the meaning of the source text and exhibits improved fluency in the target language; (2) ``\textit{Translation 2 fixes errors that were present in Translation 1}'' examines whether the output translation resolves errors present in the original translation. Additionally, we provide a free-form text box alongside each example for any additional feedback or suggestions.

\subsection{Feedback from Annotators}
\label{sec:human_eval_comments}
In our survey, we provide a text box for each example to collect additional feedback or suggestions from the annotators. We present the feedback per language pair in Table \ref{tab:human_eval_comments}. Main reasons for preferring the initial translation over the output translation from our fine-tuned model are that although the output translation may be grammatically and syntactically more precise, the initial translation often better preserves the meaning of the source sentence. Additionally, some annotators note that the initial version is more specific and understandable in some cases.

In contrast, annotators comment that they prefer the output translation from the fine-tuned model over the initial translation because (1) It fixes all the errors that were present in the initial translation; (2) It explains in the context of the target language; (3) It fits well to the actual use of the target language and flows better. All of the comments indicate that fine-tuning error annotations help make the translation more natural.

\input{tables/mqm_type}
\input{tables/error_annotation}
\input{tables/finetune_data}
\clearpage

\input{tables/llama-7}
\input{tables/llama-13}
\input{tables/llama-chat}
\clearpage

\input{tables/noerror_result}
\input{tables/fine_grained}
\input{tables/translate_from_scratch}
\clearpage

\input{tables/qualitative_zhen}
\input{tables/qualitative_ende}
\input{tables/qualitative_enru}
\clearpage

\input{figures/human_intro}
\input{figures/human_example}
\input{tables/human_eval_comments}

\end{document}

%% file: xling-write.tex
\definecolor{light green}{rgb}{0.684,0.836,0.227}

\newcommand{\ignore}[1]{}

%% file: pages/introduction.tex
\section{Introduction}

Machine Translation (MT) remains one of the last NLP tasks where large language models (LLMs) have not yet replaced dedicated supervised systems.
LLMs such as ChatGPT \cite{ouyang2022training} started outperforming commercial MT systems very recently \cite{vilar-etal-2023-prompting,hendy2023good,jiao-etal-2023-chatgpt}. However, supervised models continue to outperform LLMs in numerous language pairs \citep{zhu-etal-2023-multilingual,kocmi-etal-2023-findings}, and the performance of LLMs remains uneven, exhibiting significant variation across  models, languages, and translation directions \citep{bawden-yvon-2023-investigating,zhu-etal-2023-multilingual}. This suggests that LLMs and supervised systems possess  complementary strengths, and that combining them should offer some benefits.

\input{figures/introduction}

In this work, we propose to leverage LLM's text rewriting abilities \cite{brown2020language,reif-etal-2022-recipe,raheja-etal-2023-coedit, alves2024tower} to improve MT outputs given error annotations. \textit{If we provide an LLM with a source sentence, a MT translation of arbitrary origin, and some feedback on the quality of the MT} (Figure~\ref{fig:feedback}), \textit{can we reliably improve the quality of the MT? }
This approach can be seen as revisiting the task of MT post-editing \citep{knight-chander-1994-automated,simard-etal-2007-statistical} in the light of recent work highlighting LLMs' ability to refine its own outputs \citep{madaan2023selfrefine,zeng2023improving,chen2023iterative}. Indeed \citet{chen2023iterative,raunak2023leveraging,xu2023pinpoint} recently show the promise of using LLMs for improving MT via refinement.
We depart from these three papers by guiding the refinement abilities of LLMs with external feedback rather than self-generated feedback, and by post-editing outputs from arbitrary models rather than improve the LLM's own outputs only. Perhaps most importantly, while they relied exclusively on the largest closed LLMs \---\ GPT3.5 \cite{brown2020language}, GPT4 \cite{openai2023gpt4}, PaLM-2 \cite{anil2023palm} \---\ we argue that it is also worth exploring to what extent LLMs of more moderate size (e.g., 7B, 13B) can perform post-editing, as such models are less costly to train, run, and deploy in actual applications.  This leads us to explore a different set of strategies. We further work with open models facilitating reproducibility of our results and hopefully encourages others to build on this work.


We explore a range of techniques to guide LLaMA-2 models \cite{touvron2023llama} to improve MT outputs using fine-grained feedback derived from Multidimensional Quality Metric (MQM) annotations \cite{freitag-etal-2021-experts}, as shown in Figure~\ref{fig:feedback}. Following prior work on refinement, we start with evaluating the impact of such feedback when prompting LLMs in zero-shot and few-shot settings (\S \ref{sec:prompting_results}). Different from prior work, we then explore fine-tuning the LLM to advance its ability to improve translations based on the feedback provided in the prompt, in an instruction following style \citep{alpaca} (\S \ref{sec:fine_tuning}).

Through extensive experiments with three language pairs (Chinese-English, English-German, and English-Russian), we show that prompting LLMs to edit MT with feedback reliably improves translation quality as measured by automatic metrics, particularly in the few shot settings where the LLaMA-2 7B model achieves close peformance to the 13B version (\S \ref{sec:prompting_results}). However, the models are unable to make the most of the fine-grained feedback which performs roughly on par with generic prompts for improvement. Instruction fine-tuning shows stronger improvements on translation quality based on both automatic and human evaluation (\S \ref{sec:fine_tuning}). Our analysis reveals that prompting the fine-tuned LLMs with fine-grained feedback not only helps fix the errors highlighted in the prompt (\S \ref{sec:analysis}), but also leads to more natural outputs.

%% file: figures/introduction.tex
\begin{figure}
    \centering
    \includegraphics[width=\linewidth]{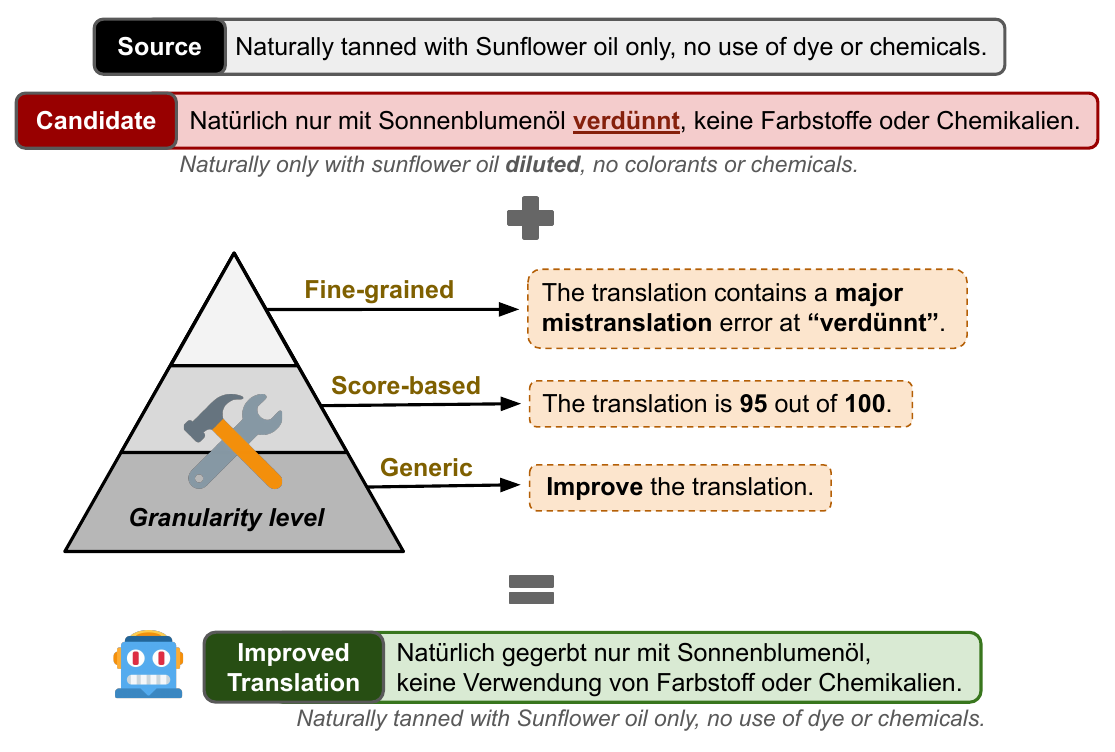}
    \caption{Guiding LLMs with external feedback enhances MT post-editing capabilities. We categorize feedback into different granularity: Generic, Score-based, and Fine-grained. Fine-grained feedback is annotated either by humans or automatic evaluation tools.}
    \label{fig:feedback}
\end{figure}

%% file: pages/related_work.tex
\section{Related Work}


\paragraph{MT Error Annotation.} An increasing body of work seeks to evaluate MT by providing actionable feedback rather than a single score aggregating diverse dimensions of quality. \citet{freitag-etal-2021-experts} introduce an evaluation methodology based on the multi-dimensional human evaluation (MQM) framework \cite{lommel-etal-2014-multidimensional} to guide human annotators in identifying spans of translated text that are errors, labeling their types and severity level using a rich taxonomy.
 Their work inspired automatic approaches to error annotation, building on existing work on automatic evaluation of text generation \cite{sellam-etal-2020-bleurt,fu2023gptscore}. These include generating a scalar score to represent MT quality as a whole \cite{xu2023pinpoint,fu2023gptscore,fernandes2023devil}, and more nuanced methods that detail error severity \cite{kocmi2023large}, error span, and type \cite{kocmi2023gembamqm}, aligning closely with human judgements \cite{liu2023geval}. Additionally, learned evaluation metrics have also emerged, pinpointing fine-grained aspects (error span, type, severity level) of MT errors \cite{guerreiro2023xcomet, xu2023pinpoint} and providing detailed error explanations \cite{xu2023instructscore}. We build on this work by comparing them using human annotated vs. machine annotated errors as feedback to refine MT outputs. 

\input{tables/related_work}

\input{tables/prompt_template}

\paragraph{MT Post-Editing.} Recognizing that translation is an iterative process, automatic post-editing originally aimed to improve an original MT provided as input together with the source text \citep{knight-chander-1994-automated,simard-etal-2007-statistical,chatterjee-etal-2018-findings}. Approaches have mirrored progress in MT, starting with statistical phrase-based models \citep{simard-etal-2007-statistical}, multi-source neural encoder-decoder models \citep{junczys-dowmunt-grundkiewicz-2016-loglinear} and non-autoregressive Transformers \citep{gu-etal-2019-levenshtein,wan-etal-2020-incorporating}. Most recent work relies on LLMs, relaxing the requirement for supervised examples of post-editing. \citet{chen2023iterative} perform refine MT outputs from a wide range of systems and languages using GPT3.5 \cite{brown2020language}, leading to a decrease of string-based quality metrics and comparable if not improved neural metrics. Human evaluation showed that this approach primarily reduces ``translationese'' in MT outputs. \citet{raunak2023leveraging} frame post-editing as  chain-of-thought \citep{kojima-etal-2023-large} and show that GPT-4 \cite{openai2023gpt4} improves COMET scores for MS Translator outputs across language pairs, particularly into English. Finally, in a contemporaneous pre-print, \citet{xu2023pinpoint} cast iterative refinement as a search process that takes as input a current MT and automatically generated MQM style error information. Using the PaLM2 LLM \cite{anil2023palm}, they show that this search improves the quality of the LLM's original translations on Chinese-English and German-English WMT tasks. Building on these encouraging results obtained with large closed models, we investigate whether smaller open LLMs can also achieve strong post-editing capabilities, which leads to explore a wider range of settings as summarized in Table \ref{tab:related_work}.

\paragraph{LLM Self-Refinement.} LLMs have been reported to ``self-correct'' an initial draft by iteratively refining it based on self-provided feedback for many tasks \citet{pan2023automatically}. Briefly, past work has focused on generation tasks including mathematical program synthesis, lexically-constrained generation, and toxicity control  \cite{welleck2022generating}, reasoning tasks \cite{paul2023refiner}, and a range of generation, math reasoning, and code optimization tasks \cite{madaan2023selfrefine}, among others. Many works focus on incorporating self-refinement to MT \cite{chen2023iterative, raunak2023leveraging, xu2023pinpoint} where given source and MT translation, LLMs generate feedback and improve upon it. In the same vein, we study MT refinement with an LLM, but incorporate error annotations from various source as feedback to refine MT outputs.

%% file: tables/related_work.tex
\begin{table}[!htp]
\centering
\resizebox{\linewidth}{!}{%
\renewcommand{\arraystretch}{1.3}
    \begin{tabular}{l l l l}
    \toprule
    \textbf{{\Large Paper}} & \textbf{{\Large Model}} & \textbf{{\Large Feedback}} & \textbf{{\Large Prompting}} \\ \midrule
    {\Large \citet{chen2023iterative}} & {\Large ChatGPT} & {\Large Self-generated} & {\Large Zero-shot} \\
    {\Large \citet{raunak2023leveraging}} & {\Large GPT-4} & {\Large Self-generated} & {\Large Zero-shot} \\
    {\Large \citet{xu2023pinpoint}} & {\Large PaLM} & {\Large Self-generated} & {\Large Zero-shot} \\
    \textbf{{\Large Ours}} & {\Large LLaMA-2} & {\Large External} & {\Large Zero-, Few-shot, Fine-tune} \\
    \bottomrule
    \end{tabular}
}
\caption{Smaller models lead us to explore a wider range of settings for post-editing with LLMs.}
\label{tab:related_work}
\end{table}

%% file: tables/prompt_template.tex
\definecolor{light gray}{rgb}{0.898, 0.902, 0.906}
\definecolor{light yellow}{rgb}{1.0, 1.0, 0.726}
\definecolor{light orange}{rgb}{0.976, 0.836, 0.625}

\begin{table*}[!htp]
\centering
\resizebox{\textwidth}{!}{%
    \begin{tabular}{l l}
    \toprule
    \textbf{Category} & \textbf{Prompt} \\ \midrule
    
    \multirow{4}{*}{\textbf{Generic}} & Improve the translation from English to German without any explanation. \\
    & English: \textit{The newer items are bagged only.} \\
    & German: \textit{Neue Gegenstände werden nur mit Gepäck versehen.} \\
    & Improved German: \\ \midrule

    \multirow{4}{*}{\textbf{Score}} & Improve the translation from English to German without any explanation. \fcolorbox{white}{light orange}{This translation is scored 85 out of 100.} \\
    & English: \textit{The newer items are bagged only.} \\
    & German: \textit{Neue Gegenstände werden nur mit Gepäck versehen.} \\
    & Improved German: \\ \midrule

    \multirow{5}{*}{\textbf{Fine-grained}} & Improve the translation from English to German \fcolorbox{white}{light orange}{based on the identified errors} without any explanation. \\
    & \fcolorbox{white}{light orange}{(1) There is a major mistranslation error at ``mit Gepäck versehen''.} \\
    & English: \textit{The newer items are bagged only.} \\
    & German: \textit{Neue Gegenstände werden nur mit Gepäck versehen.} \\
    & Improved German: \\
    
    \bottomrule
    \end{tabular}
}
\caption{Exemplar prompt template of English-German language pair used for prompting experiments. The part highlighted in \fcolorbox{white}{light orange}{orange} is the added component from the \textbf{Generic} prompt accordingly to each feedback category.}
\label{tab:prompt_template}
\end{table*}

%% file: pages/method.tex
\section{Method}

We consider two strategies for guiding language models to edit MT error annotations: prompting and fine-tuning with instructions. 

\subsection{Prompting}

We consider zero-shot and few-shot prompting. The specific prompt templates used for each feedback level are outlined in Table \ref{tab:prompt_template}, and provide a source text, a MT output and depending on the condition some feedback on the quality of the MT. We opt to construct our prompt templates in English, rather than the target language, as they have shown better performance \cite{lin-etal-2022-shot}, likely due to the greater prevalence of English in the pre-training data \cite{ahuja2023mega}. 



Our study encompasses the following forms of feedback for each model, as illustated in Table \ref{tab:prompt_template}:

\begin{itemize}[leftmargin=*, itemsep=7pt, parsep=0pt]
 \item \textbf{Generic}: The model is prompted to improve the initial translation without any specific external feedback.
 \item \textbf{Score}: A single scalar MQM score\footnote{MQM scores are derived automatically from the identified error spans and their categories \cite{fernandes2023devil}, based on a weighting scheme illustrated in Appendix Table \ref{tab:mqm_scores}.}, reflecting the initial translation's overall quality, is provided to the model. We normalize the scores on a range from 0 to 100.
  \item \textbf{Fine-grained}: The model is provided with fine-grained feedback (error span, type, severity level) in the MQM style. 

\end{itemize}

For the \textbf{Fine-grained} condition, we consider three distinct sources of error annotation:

\begin{itemize}[leftmargin=*, itemsep=7pt, parsep=0pt]
 \item \textbf{MQM}: human annotation from the MQM WMT22 dataset \cite{kocmi-etal-2022-findings}.
 \item \textbf{InstructScore}: automatic annotation by InstructScore \cite{xu2023instructscore}, an explainable text generation evaluation metric, which fine-tunes LLaMA \cite{touvron2023llama} to predict MQM style fine-grained error annotations. This metric only supports Chinese-English.
  \item \textbf{xCOMET}: automatic annotation by xCOMET \cite{guerreiro2023xcomet}, an automatic evaluation and quality estimation tool, which fine-tunes XLM-RoBERTa \cite{conneau2020unsupervised} to predict both MQM and Direct Assessment \cite{graham-etal-2013-continuous} annotations of MT quality.
\end{itemize}
The three methods use different severity level ranges, and xCOMET does not provide error type information. See Appendix \ref{sec:fine_grained} for further details.

\subsection{Fine-tuning}



In the fine-tuning case, we focus on two types of feedback: generic and fine-grained feedback, to establish the ability of fine-tuning to guide LLMs for post-editing. First, generic and fine-grained feedback consistently shows better performance compared to the score-based baseline. Second, fine-grained feedback uses human annotation thus disentangling error annotation errors from post-editing errors. We leave the exploration of automatically generated feedback to future work.

For fine-grained feedback, we explore two fine-tuning settings: (1) \textbf{Bilingual}, where we individually fine-tune for each language pair and (2) \textbf{Multilingual}, where we combine three language pairs to fine-tune a single model. We construct fine-tuning datasets from two sources of MT human-annotated with errors:  MQM \cite{freitag-etal-2021-experts} and DEMETR \cite{karpinska-etal-2022-demetr}. DEMETR provides MT error annotations in 10 source languages into English direction. Therefore, we use De-En from DEMETR as En-De pair and Ru-En as En-Ru. We reformulate all annotations in an instruction-following style (see Appendix Table~\ref{tab:finetune_data} for examples). The fine-tuning data statistics are summarized in Table \ref{tab:data_stat}. We automatically filter out instances that share identical source or target sentences with those in the test set to ensure a clean train/test separation.



%% file: pages/experimental_setup.tex
\section{Experimental Setup}

\subsection{Datasets}
\label{sec:dataset}
\paragraph{Data.} We experiment with WMT-22 General machine translation task submissions \cite{kocmi-etal-2022-findings} annotated with MQM dimensions\footnote{\url{https://github.com/google/wmt-mqm-human-evaluation}}. We focus on three language pairs: Chinese-English (zh-en), English-German (en-de), and English-Russian (en-ru). We evaluate on 1,000 WMT data instances for each language pair. Each sample contains one error span of average length ranging from 9 for En-Ru to 13 for Zh-En. Adequacy errors and minor errors dominate across languages. See Appendix~\ref{ref:mqm_dataset_details} for further details.


\input{tables/data_stat}

\paragraph{Error Annotations.} In addition to the manual error annotations described above, we obtain automatic annotations of the same data using InstructScore and xCOMET\footnote{We ensure that our data is not in their training set: InstructScore is trained on self-generated dataset from GPT-4 \cite{openai2023gpt4} and xCOMET is trained on MQM annotations but excluded the WMT-22 General MT task submissions, which they also reserved for testing.}.

To assess how much these different annotations agree with each other, we compute the overlap frequency for each pair of annotation method on a random sample of 200 test cases per language pairs. The overlap frequency  measures how often error spans match across two sources of annotations. We observe that the overlap frequency between MQM and xCOMET is 33/200 for En-De and 42/200 for En-Ru. Notably, for Zh-En pair, xCOMET and InstructScore show the highest concordance (51/200), while overlaps with MQM are lower (24/200 with xCOMET and 25/200 with InstructScore). This discrepancy underscores that the automatic annotations are far from perfect. We will test whether they can nevertheless be useful.

\input{figures/main_results}

\subsection{Metrics}

We report the traditional BLEU metric \cite{bleu} with exponential smoothing as implemented in the \texttt{sacrebleu} toolkit \cite{post-2018-call}, the Translation Edit Rate (TER) \cite{snover-etal-2006-study} which is the minimum number of edits needed to change a hypothesis so that it exactly matches one of the references, normalized by the average length of the references, and a modern neural metric, the reference-based COMET$_{\mathrm{DA}}$ score \cite{rei-etal-2020-comet}. Scores for all these metrics are reported in the 0-1 range.

\subsection{Models}

We employ the widely-used open-source LLM LLaMA-2 \cite{touvron2023llama}, experimenting with the 7B and 13B variants.\footnote{As a sanity check, we prompted the LLaMA models to translate our WMT-22 test set. The resulting translation quality (Appendix~\ref{sec:translate_from_scratch}) suggests that WMT-22 was not included in pre-training, and is therefore a valid test set.}

\paragraph{Prompting settings.} We set the temperature to 0 for greedy decoding throughout all experiments \cite{xu2023instructscore}. Through this, we ensure to reduce sampling variations of getting inconsistent generations. For 10-shot prompting, the in-context examples are chosen randomly.

\paragraph{Fine-tuning settings.}   We adopt QLoRA \cite{dettmers2023qlora}, quantized version of LoRA \cite{hu2021lora}, for parameter-efficient fine-tuning. For LoRA configs, we set the LoRA rank to 16, scaling parameter to 32, and dropout probability for layers at 0.05. We fine-tune all of the available training parameters, which is approximately 0.16B (4.4\%) of the total parameters. We use the Adam optimizer with an initial learning rate to 2e-4,  batch size of 2, gradient accumulation over 4 steps, with a warmup phase of 20 steps. We train over 5 epochs, evaluating the model's performance on 200 MQM validation set instances at the end of each epoch. We implement early stopping to halt the fine-tuning process if there is no improvement in the model performance for 16 consecutive steps.

%% file: tables/data_stat.tex
\begin{table}[H]
\centering
\resizebox{\linewidth}{!}{%
    \begin{tabular}{l c c c}
    \toprule
    \textbf{Language pair} & \textbf{\# of train} & \textbf{\# of dev} & \textbf{\# of test} \\ \midrule
    \textbf{Zh-En} & 22,373 / 3,200 & 200 & 1,000 \\
    \textbf{En-De} & 13,215 / 3,200 & 200 & 1,000 \\ 
    \textbf{En-Ru} & 19,450 / 3,200 & 200 & 1,000 \\
    \bottomrule
    \end{tabular}
}
\caption{Dataset statistics for fine-tuning instruction datasets. We use DEMETR as train set and split the MQM dataset into train, validation, and test sets. For \textbf{\# of train} column, we represent as \# of train (MQM) / \# of train (DEMETR).}
\label{tab:data_stat}
\end{table}


%% file: figures/main_results.tex
\begin{figure*}[htbp]
    \centering
    \includegraphics[width=\textwidth]{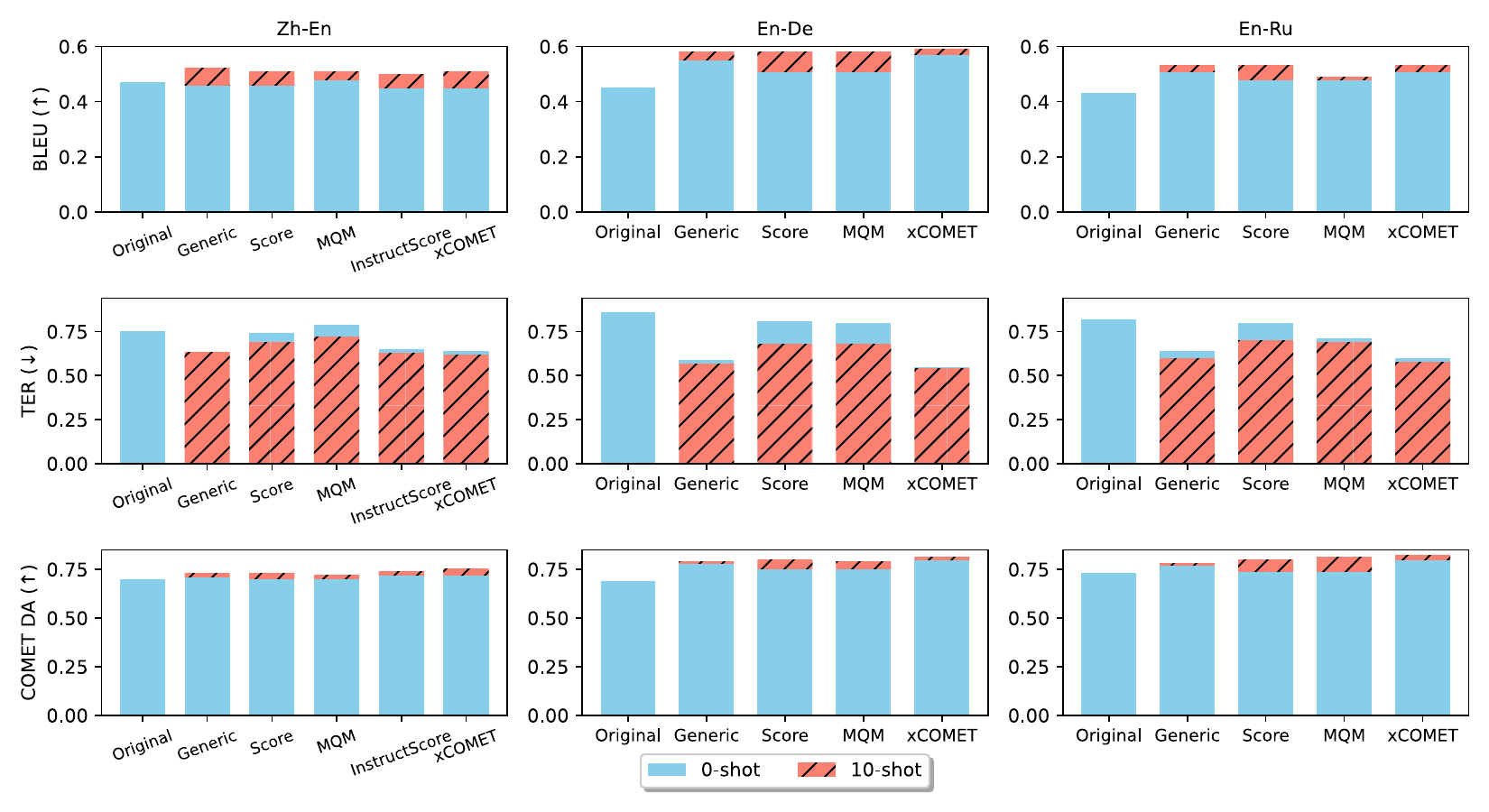}
    \caption{Zero- and 10-shot prompting results for LLaMA-2 7B. \textit{Top}: BLEU scores for Chinese-English (Zh-En), English-German (En-De), English-Russian (En-Ru) pairs; \textit{Middle}: Translation Edit Rate (TER) where zero-shot results show the amount increased compared to that of 10-shot; \textit{Bottom}: COMET$_{\mathrm{DA}}$ scores. Note that we only report the supporting language pair (zh-en) results for InstructScore. Numerical results for both 7B and 13B are in Appendix \ref{sec:llama-2-original}.} 
    \label{fig:main_results}
\end{figure*}

%% file: pages/prompting_results.tex
\section{Prompting Results}
\label{sec:prompting_results}


Figure \ref{fig:main_results} shows the zero- and 10-shot prompting performance of LLaMA-2 7B across three language pairs. The complete results in table form for both LLaMA-2 7B and 13B can be found in Appendix \ref{sec:additional_experiment}.

\paragraph{Zero-Shot.} For all language pairs, we observe a \textit{marginal improvement when post-editing with any form of feedback in zero-shot settings}, with small increases in BLEU COMET$_{\mathrm{DA}}$ scores, along with reduced TER. Although the score differences between the original and post-edited MT can be small, they are statistically significant ($p \leq 0.05$) for all cases. One exception is Zh-En pair, for which BLEU drops by 0.01 to 0.02 points after integrating feedback other than MQM.

\paragraph{Few-Shot.} The improvements from zero to 10-shot prompting are shown by hashed lines in Figure \ref{fig:main_results}. The \textit{performance gap between the original and post-edited MT widens with few-shot learning.} We examine a consistent gain in both BLEU and COMET$_{\mathrm{DA}}$ scores, which represent the overall MT quality. The average gain across language pairs is +0.04 BLEU (on a 0-1 scale) and +0.03 for COMET$_{\mathrm{DA}}$. TER, which measures the remaining amount of edits to be made also shows -0.03 point improvement for Zh-En, -0.06 point for En-De, and -0.04 point for En-Ru. 

\paragraph{7B vs 13B.} The 13B model unsurprisingly achieve higher BLEU and COMET$_{\mathrm{DA}}$ and lower TER compared to the 7B model in zero-shot settings. However, this performance gap narrows down with the increase in number of few-shot examples. This trend suggests that \textit{few-shot learning helps bridge the performance gap between model sizes} for MT post-editing. We report comprehensive results on LLaMA-2 13B in Appendix \ref{sec:additional_experiment}.

\paragraph{Feedback Granularity.} We categorize external feedback into three granularity levels: generic, score-based, and fine-grained error annotation. Fine-grained feedback is further divided into human-annotated (MQM) and automatically detected by metrics (xCOMET, InstructScore). We observe that differences in the automatic metrics across different types of feedback are small. Providing fine-grained feedback on errors has limited benefits over a generic feedback while score-based feedback shows to have the least improvement in the MT output. Overall, the performance difference between various granularity of feedback is more evident for zero-shot setting while increasing to 10-shot prompting paints a different picture.

\textit{For 10-shot prompting, most forms of our tested feedback, regardless of granularity, converge to a similar performance.} However, while the two MT quality metrics, BLEU and COMET$_\mathrm{DA}$ remains similar for different forms of feedback, there is a clear difference for TER. When providing generic feedback or automatic annotations from xCOMET, TER decreases by approximately 0.15 points for Zh-En and 0.3 points for En-De and En-Ru compared to the original baseline. Score-based feedback remains to show the least increase in performance, but they also decrease 0.1 points for Zh-En and 0.2 points for En-De and En-Ru, which are statistically significant. Nevertheless, prompting does not reveal a marked advantage for using certain type of feedback for post-editing.




\input{tables/fine_tuning}
\input{figures/error_analysis}


%% file: tables/fine_tuning.tex
\begin{table}[htbp]
\centering
\resizebox{\linewidth}{!}{%
\renewcommand{\arraystretch}{1.1}
\begin{tabular}{l l l l l}
    \toprule
    \textbf{{\large Language}} & \textbf{{\large Type}} & \textbf{{\large BLEU (↑)}} & \textbf{{\large TER (↓)}} & \textbf{{\large COMET (↑)}} \\ \midrule
    
    \multirow{6}{*}{\textbf{{\large Zh-En}}} & {\large Original} & 0.47 & 0.75 & 0.70 \\
    & {\large prompt ($k$=0)} & {\large 0.48} & {\large 0.72} & {\large 0.70} \\
    & {\large prompt ($k$=10)} & {\large 0.51} & {\large 0.65} & {\large 0.72} \\
    & {\large FT (Generic)} & {\large 0.47} & {\large 0.71} & {\large 0.72} \\
    & {\large FT (Bi)} & \textbf{{\large 0.53}}$^\dagger$ & {\large 0.63}$^\dagger$ & \textbf{{\large 0.76}}$^\dagger$ \\
    & {\large FT (Multi)} & \textbf{{\large 0.53}}$^\dagger$ & \textbf{{\large 0.61}}$^\dagger$ & \textbf{{\large 0.76}}$^\dagger$ \\ \midrule 
    
    \multirow{6}{*}{\textbf{{\large En-De}}} & {\large Original} & {\large 0.45} & {\large 0.86} & {\large 0.69} \\
    & {\large prompt ($k$=0)} & {\large 0.51} & {\large 0.68} & {\large 0.75} \\
    & {\large prompt ($k$=10)} & {\large 0.58} & {\large 0.56} & {\large 0.79} \\
    & {\large FT (Generic)} & {\large 0.52} & {\large 0.62} & {\large 0.74} \\
    & {\large FT (Bi)} & {\large 0.56}$^\dagger$ & {\large 0.58}$^\dagger$ & {\large 0.79}$^\dagger$ \\
    & {\large FT (Multi)} & \textbf{{\large 0.59}}$^\dagger$ & \textbf{{\large 0.55}}$^\dagger$ & {\large 0.79}$^\dagger$ \\ \midrule

    \multirow{6}{*}{\textbf{{\large En-Ru}}} & {\large Original} & {\large 0.43} & {\large 0.82} & {\large 0.73} \\
    & {\large prompt ($k$=0)} & {\large 0.48} & {\large 0.69} & {\large 0.74} \\
    & {\large prompt ($k$=10)} & {\large 0.49} & {\large 0.67} & {\large 0.79} \\
    & {\large FT (Generic)} & {\large 0.46} & {\large 0.67} & {\large 0.74} \\
    & {\large FT (Bi)} & {\large 0.51}$^\dagger$ & {\large 0.65}$^\dagger$ & \textbf{{\large 0.80}}$^\dagger$ \\
    & {\large FT (Multi)} & \textbf{{\large 0.52}}$^\dagger$ & \textbf{{\large 0.63}}$^\dagger$ & \textbf{{\large 0.80}}$^\dagger$ \\
    
    \bottomrule
    \end{tabular}
}
\caption{Fine-tuning (FT) results for LLaMA-2 7B. \textbf{prompt ($k$=0)} and \textbf{prompt ($k$=10)} indicate the zero- and 10-shot prompting results of LLaMA-2 7B respectively. \textbf{FT (Generic)}: Fine-tuning with generic feedback; \textbf{FT (Bi)}: Fine-tuning with fine-grained feedback in bilingual setting, where models are individually fine-tuned for each language pair; \textbf{FT (Multi)}: Fine-tuning with fine-grained feedback in multilingual setting, combine 3 language pairs to fine-tune a single model. We test the statistically significance of improvements over the best prompting baseline and $\dagger$ marks results with $p$-value $\leq 0.05$.}
\label{tab:fine_tuning}
\end{table}

%% file: figures/error_analysis.tex
\begin{figure*}[!htp]
    \centering
    \includegraphics[width=\textwidth]{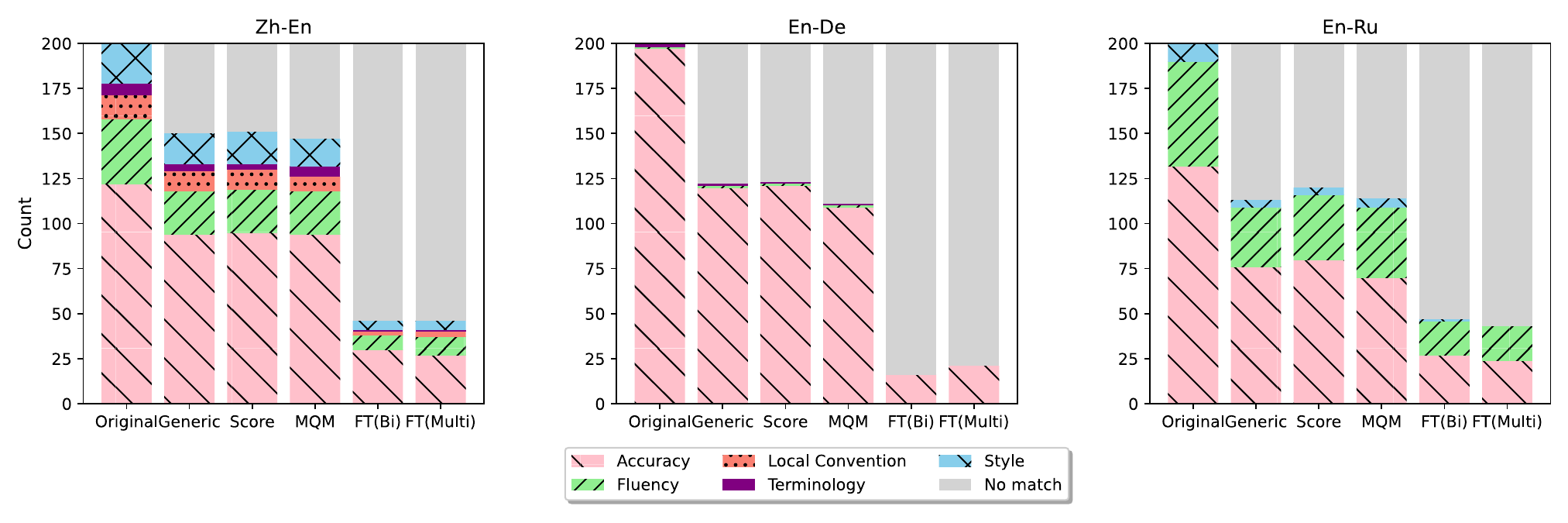}
    \caption{Error analysis for LLaMA-2 7B. We observe how much each error type is resolved by integrating external feedback during post-editing. We classify an error as `\textit{No match}' if the output translation does not contain the specific error span. Across all language pairs, fine-tuning best addresses the errors present in the initial translation. \textbf{FT (Bi)}: fine-tuning in bilingual setting; \textbf{FT (Multi)}: fine-tuning in multilingual setting. We do not include InstructScore or xCOMET as InstructScore annotates more than 1 error spans, making it difficult for fair comparison and xCOMET does not output error type information.}
    \label{fig:error_analysis}
\end{figure*}

%% file: pages/ft_results.tex
\section{Fine-Tuning Results}
\label{sec:fine_tuning}

\subsection{Automatic Evaluation}
\label{sec:auto_eval}

We examine the effectiveness of fine-tuning error-annotated translations for MT post-editing. Table \ref{tab:fine_tuning} shows that \textit{fine-tuning with error annotated translations gives an extra boost in the performance across all metrics.}

\paragraph{Original vs Fine-tuning.}
We compare the fine-tuning results of each language pair against the original translation quality (indicated as `Original' in Table \ref{tab:fine_tuning}). Across language pairs, metrics of MT quality all increase for fine-tuning. Translation quality increases steeply by approximately +0.07 BLEU, +0.08 COMET$_\mathrm{DA}$ and -0.21 TER on average for all language pairs. The multilingual approach mostly outperforms the bilingual one, suggesting that the advantages gleaned from fine-tuning with diverse language pairs outweigh the benefits of matching the fine-tuning data language consistent to the test language pair. We observe the same trend with LLaMA-2 13B in Appendix Table \ref{tab:fine_tuning_13}: fine-tuning results improve upon the original baseline results by +0.1 BLEU, +0.08 COMET$_\mathrm{DA}$ and -0.25 TER points on average.

\paragraph{Prompting vs Fine-tuning.}
Next, we examine fine-tuning evaluation compared to the zero- and 10-shot prompting results, collected from either LLaMA-2 7B or 13B. Compared to zero-shot prompting, fine-tuning with error annotations always outperform across all metrics and the multilingual approach outperforms 10-shot prompting results for most of the cases.



\paragraph{Feedback granularity.}
We compare the two distinct types of feedback used for fine-tuning: generic and fine-grained feedback, denoted as `FT (Generic)' and `FT (Multi)' respectively in Table \ref{tab:fine_tuning}. While prompting experiments demonstrate no clear preference between levels of feedback granularity, fine-tuning using fine-grained feedback consistently yields superior translation quality compared to fine-tuning with generic feedback with a gap of 4 to 6 BLEU points, 3 to 8 TER, and 4 to 6 COMET. This shows that fine-tuning allows the models to take advantage of the fine-grained feedback more effectively.


As there are few error tokens overall, we first expected to see small edits from our fine-tuned model, thus small score difference. However, surprisingly, fine-tuning results overall show greater improvements, especially for TER, considering that the original MQM dataset only has one error span per sentence. Examining outputs (see Appendix \ref{sec:fine_tune_qualitative} for examples) suggests that fine-tuning not only edits the targeted error spans but also improve the overall naturalness in the target language, consistent with prior evidence that post-editing with LLMs reduces translationese effects \citep{chen2023iterative}. To further validate this hypothesis, we turn to human evaluation.



\subsection{Human Evaluation}
\label{sec:human_eval}

We ask bilingual human annotators to assess the post-edited outputs obtained by fine-tuning in the bilingual setting as it is the stronger approach based on automatic scores. We randomly select 50 instances for each language pair for annotation. Each instance is examined by 3 human annotators. For each instance of source text, original MT with MQM annotation, post-edited MT, the annotator is asked to rate on a 5-point Likert scale (1 strongly disagree to 5 strongly agree) whether the translation quality has improved, and to what extent the annotated errors are actually resolved through post-editing. Ordinal Kripendorff's alpha \cite{Krippendorff2011ComputingKA}\footnote{Kripendorff's alpha ranges from 0 to 1, where 0 means no agreement and 1 means perfect agreement.}, which measure the inter-annotator agreement is moderate for the \textit{Overall quality}: 0.527, 0.479, 0.421 for Zh-En, En-De, and En-Ru. Annotators are also given the option to provide free form comments. Refer to Appendix~\ref{sec:human_eval_details} for further details on the annotation set-up.

\input{figures/human_eval} 


As illustrated in Figure \ref{fig:human_eval}, our human evaluation results confirm that fine-tuning with error annotations enhances overall translation quality (\textit{Overall Quality}) and effectively resolves errors in the initial translation (\textit{Resolve Errors}). While this improvement is notably evident in Zh-En and En-De pair, for the En-Ru pair, approximately 40/150 annotations lean towards the \textit{Disagree} category. Some of the feedback from En-Ru annotators who choose to \textit{Disagree} state that there are cases when the output translation from the fine-tuned model is more precise in the target language, but loses some of the nuance in the source text.

Further, feedback from the annotators support our own observation that the post-editing via fine-tuning does not only fix targeted errors in the original translation but rewrites for naturalness in the target language. They comment that the fine-tuning translation ``\textit{better explains the context}'' and ``\textit{flows better in the target language}'' compared to the original translation which seems to be directly translated without consideration of the context. We list further comments in Appendix Table \ref{tab:human_eval_comments}.



%% file: figures/human_eval.tex
\begin{figure}
    \centering
    \includegraphics[width=\linewidth]{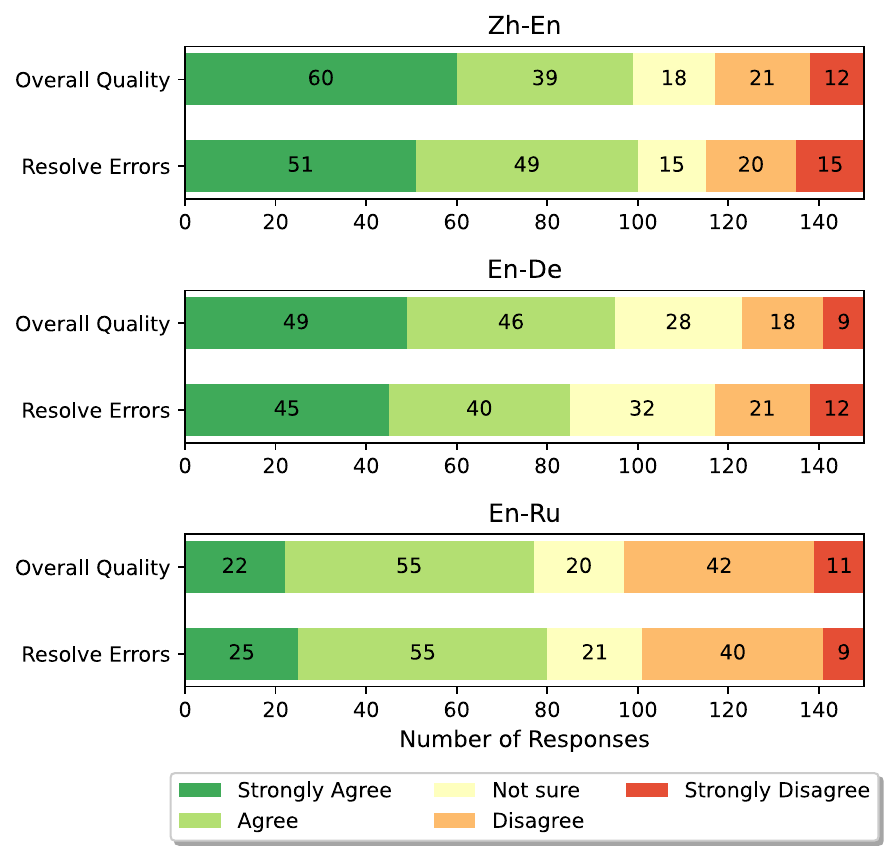}
    \caption{Human evaluation results for 3 language pairs. We collect a total of 150 annotations for each language pair. \textit{Overall Quality}: Output translation from the fine-tuned model is better than the initial translation; \textit{Resolve Errors}: Output translation resolves errors in the initial translation.}
    \label{fig:human_eval}
\end{figure}

%% file: pages/analysis.tex
\section{Analysis by MT Error Categories}
\label{sec:analysis}

Our error analysis aims to pinpoint the types of errors that are most effectively resolved through the integration of external feedback. We evaluate 200 output translations generated by prompting LLaMA-2 7B with each generic, score-based, and MQM feedback. We do not include InstructScore or xCOMET as InstructScore annotates more than 1 error spans making it difficult for fair comparison and xCOMET does not output error type information. We also compare the outputs from our custom fine-tuned models, both bilingual and multilingual version. All of the feedback is based on MQM, thus we categorize the error type as per ``Error Category'' from MQM detailed in Appendix Table \ref{tab:mqm_type}.



In Figure \ref{fig:error_analysis}, we illustrate the extent to which each error type has been resolved by incorporating external feedback. First, we check whether a span annotated as an error in the original translation matches the output after post-editing with feedback. A match increments the count for the error type associated with the span. If there is no match found, the count for the ``No match'' category is incremented. We observe that using any form of feedback (generic, score, or MQM) increases the portion of ``No match'' category compared to the original translation. However, there is no distinct trend for any specific error type; all of the errors are addressed in a balanced manner. 

Further, by incorporating the output translations from our fine-tuned model, we see a sudden leap in the ``No match'' category. This suggests that fine-tuning best fixes the targeted error span. This finding is also consistent with the conclusions from Section \ref{sec:fine_tuning}, where we noted that fine-tuning help align LLM behavior with the provided feedback.


%% file: pages/analysis_noerror.tex
\section{Post-Editing Correct Outputs}
\label{sec:analysis_noerror_outputs}

The experiments we have presented so far are focused on post-editing MT hypotheses that are known to leave room for improvement. For completeness, we present in Appendix Table \ref{tab:noerror_result} decoding results when zero-shot prompting the LLaMA-2 models to post-edit approaches to 200 WMT hypotheses labeled as ``No error'' by the WMT human annotators. 

As expected, the resulting edits lead to a small drop in automatic metrics, confirming the observation that the nature of edits goes beyond correcting errors to address more stylistic issues such as translationese. Interestingly, the larger LLaMA-2 model and the fine-grained feedback are the least prone to over-editing. We anticipate that different prompts and fine-tuning data are needed for models to jointly consider the task of editing or not, and of what edits to perform.

%% file: pages/conclusion.tex
\section{Conclusion}
We explore a range of strategies to guide LLaMA-2 models to improve MT outputs using external feedback, varying in different granularity. We demonstrate that prompting LLM to edit MT with feedback reliably enhances the overall translation quality and post-editing efforts. We further explore instruction fine-tuning LLMs with fine-grained feedback. Through automatic and human evaluation, we demonstrate that fine-tuning shows stronger improvements on enhancing the translation quality, resolving errors in the initial translation, and most notably, generating translations that are more natural (less translationese) in the target language. 

Taken together, these results clearly show that post-editing MT output does not require the largest proprietary LLM models and can be done with smaller open-source models. This opens many questions for future work to further explore how to do this well in more diverse settings, while minimizing the reliance on human annotated MT outputs which are expensive to obtain at scale. Building on LLMs fine-tuned for many translation related tasks \cite{alves2024tower} is a promising direction for encouraging transfer learning from limited amounts of annotation.

%% file: pages/limitation.tex
\section{Limitations}


We evaluate the impact of post-editing separately on MT outputs that contain one or more errors (\S \ref{sec:prompting_results}) and on MT outputs that do not contain any errors (\S \ref{sec:analysis_noerror_outputs}). This leaves open the question of how to design a workflow that takes in any MT input and automatically determines whether and how it should be post-edited, possibly selecting among different potential feedback mechanisms, which we leave to future work.

Furthermore, the fine-tuning data is in the same domain as the test data which will not always be the case in practice. While we test on diverse languages and on out-of-English and into-English directions, it remains to be seen how our findings generalize to a wider variety of languages, particularly in low-resource settings. 

Finally, our work highlights the effectiveness of using external feedback to resolve errors in translations. Although integrating external feedback is an attractive approach, the scarcity of high-quality feedback remains a significant challenge. This scarcity underscores the demand for the development of automated systems capable of generating high-quality error annotations. In regard to constraints in the currently available external feedback for MT post-editing, our study is constrained in terms of forms of feedback (generic, score, fine-grained) and language pairs. Future works can focus on incorporating automatic systems that can generate consistent, high quality feedback.



%% file: pages/acknowledgement.tex
\section{Acknowledgement}
\label{sec:analysis_noerror}

We thank the anonymous reviewers, Shramay Palta, Nishant Balepur, Calvin Bao, Yu Hou and the members of the \textsc{clip} lab at University of Maryland for their valuable suggestions and constructive feedback.

This work was supported in part by NSF Fairness in AI Grant 2147292, by the Institute for Trustworthy AI in Law and Society (TRAILS), which is supported by the National Science Foundation under Award No. 2229885, and by the Office of the Director of National Intelligence (ODNI), Intelligence Advanced Research Projects Activity (IARPA), via the HIATUS Program contract \#2022-22072200006, by NSF grant 2147292. The views and conclusions contained herein are those of the authors and should not be interpreted as necessarily representing the official policies, either expressed or implied, of ODNI, IARPA, NSF or the U.S. Government. The U.S. Government is authorized to reproduce and distribute reprints for governmental purposes notwithstanding any copyright annotation therein.

%% file: tables/severity_level.tex
\begin{table}[!h]
\centering
\resizebox{\linewidth}{!}{%
    \begin{tabular}{l c}
    \toprule
    \textbf{Metric} & \textbf{Severity level} \\ \midrule
    \textbf{MQM} & Major, Minor, No-error  \\
    \textbf{InstructScore} & Major, Minor \\
    \textbf{xCOMET} & Critical, Major, Minor \\
    \bottomrule
    \end{tabular}
}
\caption{Error severity levels supported by each metric.}
\label{tab:severity_level}
\end{table}

%% file: tables/mqm_scores.tex
\begin{table}[!h]
\centering
\resizebox{\linewidth}{!}{%
    \begin{tabular}{l c c}
    \toprule
    \textbf{Severity} & \textbf{Category} & \textbf{Weight} \\ \midrule
    \multirow{2}{*}{\textbf{Major}} & Non-translation & 25 \\
    & All others & 5 \\ \midrule
    \multirow{2}{*}{\textbf{Minor}} & Fluency/Punctuation & 0.1 \\ 
    & All others & 1 \\ \midrule
    \textbf{Neutral} & All & 0 \\
    \bottomrule
    \end{tabular}
}
\caption{MQM error weighting. Each score can range from 0 (perfect) to 25 (worst). The final score is the average over scores from all annotators.}
\label{tab:mqm_scores}
\end{table}

%% file: figures/mqm_type_distribution.tex

\begin{figure*}[htbp]
    \centering
    \begin{minipage}{0.48\linewidth}
        \includegraphics[width=\linewidth]{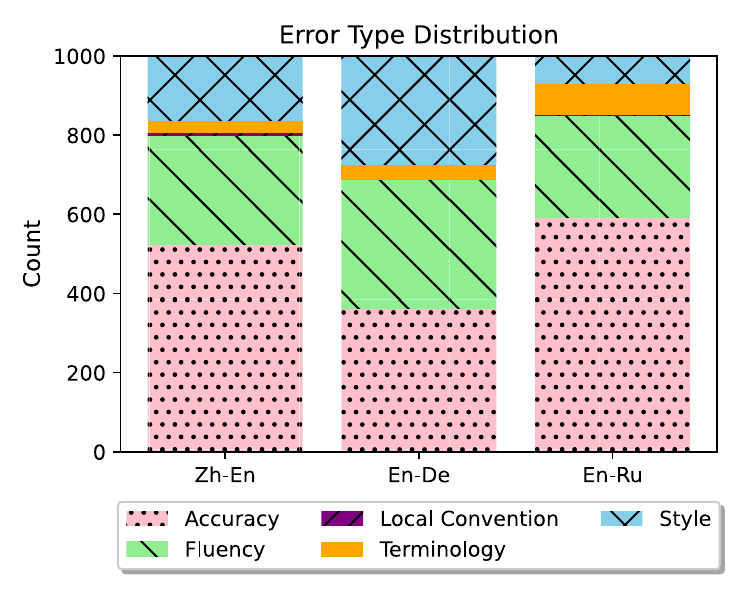}
        \caption{Error type distribution for 3 language pairs. Note that En-Ru dataset from WMT 22 General MT submissions use different names for each error type, thus conduct manual mapping.}
        \label{fig:mqm_type_distribution}
    \end{minipage}
    \hfill
    \begin{minipage}{0.48\linewidth}
        \includegraphics[width=\linewidth]{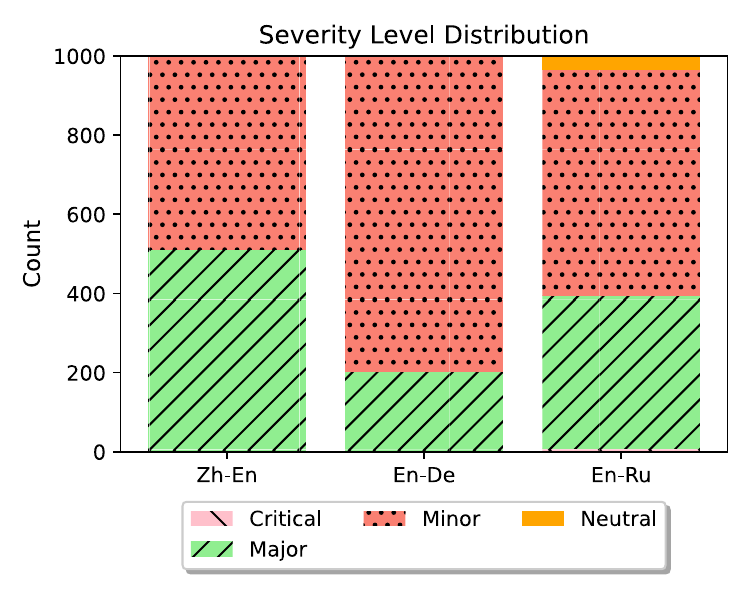}
        \caption{Severity level distribution for 3 language pairs. We note that the En-Ru dataset from WMT 22 General MT submissions use additional severity level category: ``\textit{critical}''.}
        \label{fig:mqm_severity_distribution}
    \end{minipage}
\end{figure*}

%% file: tables/fine_tuning_13.tex
\begin{table}[htbp]
\centering
\resizebox{\linewidth}{!}{%
\renewcommand{\arraystretch}{1.1}
\begin{tabular}{l l l l l}
    \toprule
    \textbf{{\large Language}} & \textbf{{\large Type}} & \textbf{{\large BLEU (↑)}} & \textbf{{\large TER (↓)}} & \textbf{{\large COMET$_\mathrm{DA}$ (↑)}} \\ \midrule

    \multirow{5}{*}{\textbf{{\large Zh-En}}} & {\large Original} & {\large 0.47} & {\large 0.75} & {\large 0.70} \\
    & {\large prompt ($k$=0)} & {\large 0.50} & {\large 0.74} & {\large 0.71} \\
    & {\large prompt ($k$=10)} & {\large 0.50} & {\large 0.61} & {\large 0.73} \\
    & {\large FT (Bi)} & {\large 0.51}$^\dagger$ & {\large 0.61} & \textbf{{\large 0.77}}$^\dagger$ \\
    & {\large FT (Multi)} & \textbf{{\large 0.54}}$^\dagger$ & \textbf{{\large 0.58}}$^\dagger$ & {\large 0.76}$^\dagger$ \\ \midrule 
    
    \multirow{5}{*}{\textbf{{\large En-De}}} & {\large Original} & {\large 0.45} & {\large 0.86} & {\large 0.69} \\
    & {\large prompt ($k$=0)} & {\large 0.51} & {\large 0.68} & {\large 0.75} \\
    & {\large prompt ($k$=10)} & {\large 0.58} & {\large 0.56} & {\large 0.79} \\
    & {\large FT (Bi)} & {\large 0.57} & {\large 0.55}$^\dagger$ & \textbf{{\large 0.80}}$^\dagger$ \\
    & {\large FT (Multi)} & \textbf{{\large 0.60}}$^\dagger$ & \textbf{{\large 0.53}}$^\dagger$ & \textbf{{\large 0.80}}$^\dagger$ \\ \midrule

    \multirow{5}{*}{\textbf{{\large En-Ru}}} & {\large Original} & {\large 0.43} & {\large 0.82} & {\large 0.73} \\
    & {\large prompt ($k$=0)} & {\large 0.44} & {\large 0.80} & {\large 0.73} \\
    & {\large prompt ($k$=10)} & {\large 0.53} & {\large 0.56} & {\large 0.80} \\
    & {\large FT (Bi)} & {\large 0.53} & {\large 0.57} & {\large 0.80} \\
    & {\large FT (Multi)} & \textbf{{\large 0.54}}$^\dagger$ & {\large 0.56}$^\dagger$ & \textbf{{\large 0.81}}$^\dagger$ \\
    
    \bottomrule
    \end{tabular}
}
\caption{Fine-tuning (FT) results for LLaMA-2 13B. \textbf{prompt ($k$=0)} and \textbf{prompt ($k$=10)} indicate the zero- and 10-shot prompting results of LLaMA-2 13B respectively. \textbf{FT (Bi)}: Fine-tuning in bilingual setting, where models are individually fine-tuned for each language pair; \textbf{FT (Multi)}: Fine-tuning in multilingual setting, combine 3 language pairs to fine-tune a single model. We test the statistically significance of improvements over the best prompting baseline and $\dagger$ marks results with $p$-value $\leq 0.05$.}
\label{tab:fine_tuning_13}
\end{table}

%% file: tables/mqm_type.tex
\begin{table*}[!htp]
\centering
\resizebox{\textwidth}{!}{%
    \begin{tabular}{l l l}
    \toprule
    \textbf{Error Category} & \textbf{Sub Category} & \textbf{Description}\\ \midrule
    
    \textbf{Accuracy} & Addition & Translation includes information not present in the source.\\
    & Omission &  Translation has missing content from the source.\\
    & Mistranslation & Target content does not accurately represent the source content.\\
    & Untranslated text & Source text has been left untranslated. \\ \midrule

    \textbf{Fluency} & Character Encoding & Characters are garbled due to incorrect application of an encoding.\\
    & Grammar & Problems with grammar or syntax of text, other than orthography. \\
    & Inconsistency & Internal inconsistency (not related to terminology). \\
    & Punctuation & Incorrect punctuation (for locale or style). \\
    & Register & Wrong grammatical register (eg. informal pronouns or verb forms) \\ 
    & Spelling & Incorrect spelling or capitalization. \\ \midrule

    \textbf{Local convention} & Address format & Wrong format for addresses. \\
    & Currency format & Wrong format for currency. \\
    & Date format & Wrong format for dates. \\
    & Name format & Wrong format for names. \\
    & Telephone format & Wrong format for telephone numbers. \\
    & Time format & Wrong format for time expressions. \\ \midrule

    \textbf{Terminology} & Inappropriate for context & Terminology is non-standard or does not fit context. \\
    & Inconsistent use & Terminology is used inconsistently. \\ \midrule
    
    \textbf{Style} & Awkward & Translation has stylistic problems. \\ \midrule

    \textbf{Source error} &  & Any error in the source. \\ \midrule
    \textbf{Non-translation} &  & Impossible to reliably characterize distinct errors. \\ \midrule
    \textbf{Other} &  & Any other issues. \\ 
    \bottomrule
    \end{tabular}
}
\caption{MQM hierarchy \cite{freitag-etal-2021-experts}.}
\label{tab:mqm_type}
\end{table*}

%% file: tables/error_annotation.tex
\begin{CJK*}{UTF8}{gbsn}

\begin{table*}[!htp]
\centering
\resizebox{\textwidth}{!}{%
    \begin{tabular}{l l}
    \toprule
    \textbf{Source} & 现如今绝大多数遇难者的老父母均已谢世，遗孤们也已长大成家就业。 \\ \midrule
    \textbf{Candidate} & Nowadays, the vast majority of the victims' elderly parents have died, and the orphans have grown into family employment. \\ \midrule
    \textbf{Reference} & Now the old parents of the vast majority of the victims have passed away, and the orphans have also grown up, started working and got married. \\ \midrule
    
    \multirow{3}{*}{\textbf{MQM}} & Error span: Nowadays \\
    & Error type: Accuracy/Mistranslation \\ 
    & Severity: Major \\ \midrule
    
    \multirow{3}{*}{\textbf{InstructScore}} & Error span: [family employment, Nowadays] \\ 
    & Error type: [Incorrect translation is missing content from the correct translation, Incorrect translation has stylistic problems] \\
    & Severity: [Major, Major] \\ \midrule
   
    \multirow{2}{*}{\textbf{xCOMET}} & Error span: [die, have grown into family employment] \\ 
    & Severity: [Major, Major]\\
    \bottomrule
    \end{tabular}
}
\caption{Chinese-English error annotation examples from three sources: MQM dataset, xCOMET, and InstructScore. We input source, candidate translation, and reference sentence to xCOMET and InstructScore. xCOMET does not output error type in their annotation. Both xCOMET and InstructScore returns a list if they detect more than one errors in the candidate translation.}
\label{tab:error_annotate}
\end{table*}

\end{CJK*}

%% file: tables/finetune_data.tex
\definecolor{light gray}{rgb}{0.898, 0.902, 0.906}
\definecolor{light green}{rgb}{0.827, 0.969, 0.773}
\definecolor{light blue}{rgb}{0.773, 0.827, 0.969}

\begin{table*}[htbp]
\centering
\resizebox{\textwidth}{!}{%
    \begin{tabular}{l}
    \toprule
    \textbf{Instruction} \\ \midrule
    
    \fcolorbox{white}{light gray}{\#\#\# \textbf{English}:} Memorial meetings were organised at the residence of Sam Stafford, one of the agitators who died, and a playground in Guwahati, \\ with attendees resolving to once again to intensify the stir against the Citizenship (Amendment) Act.\textcolor{gray}{\textbackslash n} \\
    \fcolorbox{white}{light gray}{\#\#\# \textbf{German}:} Gedenkmälerversammlungen wurden in der Residenz von Sam Stafford, einem der gestorbenen Agitatoren, und einem Spielplatz \\ in Guwahati organisiert, wobei die Teilnehmer sich entschlossen hatten, den Aufruhr gegen das Gesetz über die Staatsbürgerschaft (Änderung) \\ erneut zu intensivieren.\textcolor{gray}{\textbackslash n} \\
    \fcolorbox{white}{light blue}{\#\#\# Errors:} There is a minor accuracy/mistranslation error at ``Gedenkmälerversammlungen''.\textcolor{gray}{\textbackslash n\textbackslash n} \\
    \fcolorbox{white}{light green}{\#\#\# Improved \textbf{German}:} Gedenkveranstaltungen fanden am Wohnsitz von Sam Stafford, einem der getöteten Aktivisten, sowie auf einem Schulhof \\ in Guwahati statt, und die Teilnehmer beschlossen, noch einmal den Protest gegen den CAA zu verstärken. \\
    \bottomrule
    \end{tabular}
}
\caption{Example of fine-tuning instructions dataset reformulated from English-German error annotated translations. Texts in \textbf{bold} represent the placeholders for source and target language. We give fine-grained feedback after \fcolorbox{white}{light blue}{\#\#\# Errors:}.}
\label{tab:finetune_data}
\end{table*}

%% file: tables/llama-7.tex
\definecolor{light green}{rgb}{0.645, 0.898, 0.711}
\definecolor{light red}{rgb}{0.992, 0.678, 0.678}

\begin{table*}
\centering
\resizebox{\textwidth}{!}{%
    \begin{tabular}{l c c c c c c c}
    \toprule
    \multirow{2}{*}{\textbf{Language}} & \multirow{2}{*}{\textbf{Shots}} & \multicolumn{6}{c}{\textbf{BLEU (↑)} / \textbf{TER (↓)} / \textbf{COMET$_\mathrm{DA}$ (↑)}} \\
    & & Original & Generic & Score & MQM & InstructScore & xCOMET \\ \midrule
    
    \multirow{2}{*}{\textbf{Zh-En}} & 0 & \multirow{2}{*}{0.47 / 0.75 / 0.70} & \fcolorbox{white}{light red}{0.45}$^\dagger$ / 0.63$^\dagger$ / 0.71$^\dagger$ & \fcolorbox{white}{light red}{0.46}$^\dagger$ / 0.69$^\dagger$ / 0.7$^\dagger$ & 0.48$^\dagger$ / 0.72$^\dagger$ / 0.70$^\dagger$ & \fcolorbox{white}{light red}{0.45}$^\dagger$ / 0.63$^\dagger$ / 0.72$^\dagger$ & \fcolorbox{white}{light red}{0.45}$^\dagger$ / 0.62$^\dagger$ / 0.72$^\dagger$ \\
    
    & 10 & & \fcolorbox{white}{light green}{0.51}$^\dagger$ / 0.63$^\dagger$ / 0.73$^\dagger$ & \fcolorbox{white}{light green}{0.51}$^\dagger$ / 0.64$^\dagger$ / 0.73$^\dagger$ & \fcolorbox{white}{light green}{0.51}$^\dagger$ / 0.65$^\dagger$ / 0.72$^\dagger$ & 0.50$^\dagger$ / 0.61$^\dagger$ / 0.74$^\dagger$ & \fcolorbox{white}{light green}{0.51}$^\dagger$ / \fcolorbox{white}{light green}{0.60}$^\dagger$ / \fcolorbox{white}{light green}{0.75}$^\dagger$ \\ \midrule
    
    \multirow{2}{*}{\textbf{En-De}} & 0 & \multirow{2}{*}{0.45 / 0.86 / 0.69} & 0.55$^\dagger$ / 0.57$^\dagger$ / 0.78$^\dagger$ & 0.51$^\dagger$ / 0.68$^\dagger$ / 0.75$^\dagger$ & 0.51$^\dagger$ / 0.68$^\dagger$ / 0.75$^\dagger$ & - & \fcolorbox{white}{light green}{0.57}$^\dagger$ / \fcolorbox{white}{light green}{0.54}$^\dagger$ / 0.80$^\dagger$ \\
    
    & 10 & & 0.58$^\dagger$ / 0.55$^\dagger$ / 0.79$^\dagger$ & 0.55$^\dagger$ / 0.59$^\dagger$ / 0.80$^\dagger$ & 0.56$^\dagger$ / 0.58$^\dagger$ / 0.79$^\dagger$ & - & 0.53$^\dagger$ / 0.57$^\dagger$ / \fcolorbox{white}{light green}{0.81}$^\dagger$ \\ \midrule
    
    \multirow{2}{*}{\textbf{En-Ru}} & 0 & \multirow{2}{*}{0.43 / 0.82 / 0.73} & 0.51$^\dagger$ / 0.60$^\dagger$ / 0.77$^\dagger$ & 0.48$^\dagger$ / 0.70$^\dagger$ / 0.74$^\dagger$ & 0.48$^\dagger$ / 0.69$^\dagger$ / 0.74$^\dagger$ & - & 0.51$^\dagger$ / 0.58$^\dagger$ / 0.80$^\dagger$ \\
    
    & 10 & & 0.53$^\dagger$ / 0.56$^\dagger$ / 0.79$^\dagger$ & \fcolorbox{white}{light green}{0.53}$^\dagger$ / 0.60$^\dagger$ / 0.79$^\dagger$ & 0.49$^\dagger$ / 0.70$\ddagger$ / 0.79$^\dagger$ & - & \fcolorbox{white}{light green}{0.53}$^\dagger$ / \fcolorbox{white}{light green}{0.56}$^\dagger$ / \fcolorbox{white}{light green}{0.82}$^\dagger$ \\
    
    \bottomrule
    \end{tabular}
}
\caption{Zero- and 10-shot prompting performance of LLaMA-2 7B model. Original column measures between the source and target sentences from the original MQM dataset. Other columns represents the model performance for different types of feedback: Generic, Score, Fine-grained (MQM, xCOMET, InstructScore). \fcolorbox{white}{light green}{Green}: best performance per language pair; \fcolorbox{white}{light red}{Red}: worse performance than the original baseline. We test the statistically significance of improvements over the original and $\dagger$ marks results with $p$-value $\leq 0.05$ and $\ddagger$ marks results with $p$-value $\leq 0.1$.}
\label{tab:llama-7}
\end{table*}

%% file: tables/llama-13.tex
\definecolor{light green}{rgb}{0.645, 0.898, 0.711}
\definecolor{light red}{rgb}{0.992, 0.678, 0.678}

\begin{table*}
\centering
\resizebox{\textwidth}{!}{%
    \begin{tabular}{l c c c c c c c}
    \toprule
    \multirow{2}{*}{\textbf{Language}} & \multirow{2}{*}{\textbf{Shots}} & \multicolumn{6}{c}{\textbf{BLEU (↑)} / \textbf{TER (↓)} / \textbf{COMET$_\mathrm{DA}$ (↑)}} \\
    & & Original & Generic & Score & MQM & InstructScore & xCOMET \\ \midrule
    
    \multirow{2}{*}{\textbf{Zh-En}} & 0 & \multirow{2}{*}{0.47 / 0.75 / 0.70} & 0.50$^\dagger$ / 0.66$^\dagger$ / 0.73$^\dagger$ & 0.50$^\dagger$ / 0.72$^\dagger$ / 0.72 & 0.50$^\dagger$ / 0.74$^\dagger$ / 0.71$^\dagger$ & 0.50$^\dagger$ / 0.61$^\dagger$ / 0.75$^\dagger$ & 0.53$^\dagger$ / 0.59$^\dagger$ / \fcolorbox{white}{light green}{0.76}$^\dagger$ \\
    
    & 10 & & 0.51$^\dagger$ / 0.61$^\dagger$ / 0.74$^\dagger$ & 0.51$^\dagger$ / 0.61$^\dagger$ / 0.74$^\dagger$ & 0.50$^\dagger$ / 0.61$^\dagger$ / 0.73$^\dagger$ & 0.50$^\dagger$ / 0.61$^\dagger$ / 0.75$^\dagger$ & \fcolorbox{white}{light green}{0.54}$^\dagger$ / \fcolorbox{white}{light green}{0.58}$^\dagger$ / \fcolorbox{white}{light green}{0.76}$^\dagger$ \\ \midrule
    
    \multirow{2}{*}{\textbf{En-De}} & 0 & \multirow{2}{*}{0.45 / 0.86 / 0.69} & 0.58$^\dagger$ / 0.55$^\dagger$ / 0.80$^\dagger$ & 0.49$^\dagger$ / 0.73$^\dagger$ / 0.73$^\dagger$ & 0.48$^\dagger$ / 0.76$^\dagger$ / 0.72 & - & 0.60$^\dagger$ / 0.52$^\dagger$ / 0.81$^\dagger$ \\
    
    & 10 & & 0.58$^\dagger$ / 0.54$^\dagger$ / 0.80$^\dagger$ & 0.58$^\dagger$ / 0.55$^\dagger$ / 0.80$^\dagger$ & 0.58$^\dagger$ / 0.54$^\dagger$ / 0.80$^\dagger$ & - & \fcolorbox{white}{light green}{0.62}$^\dagger$ / \fcolorbox{white}{light green}{0.51}$^\dagger$ / \fcolorbox{white}{light green}{0.82}$^\dagger$ \\ \midrule
    
    \multirow{2}{*}{\textbf{En-Ru}} & 0 & \multirow{2}{*}{0.43 / 0.82 / 0.73} & 0.53$^\dagger$ / 0.56$^\dagger$ / 0.79$^\dagger$ & 0.46$^\dagger$ / 0.74$^\dagger$ / 0.74$^\dagger$ & 0.44$^\dagger$ / 0.80 / 0.73 & - & 0.55$^\dagger$ / 0.54$^\dagger$ / 0.83$^\dagger$ \\
    
    & 10 & & 0.54$^\dagger$ / 0.55$^\dagger$ / 0.80$^\dagger$ & 0.54$^\dagger$ / 0.56$^\dagger$ / 0.80$^\dagger$ & 0.53$^\dagger$ / 0.56$^\dagger$ / 0.80$^\dagger$ & - & \fcolorbox{white}{light green}{0.57}$^\dagger$ / \fcolorbox{white}{light green}{0.52}$^\dagger$ / \fcolorbox{white}{light green}{0.85}$^\dagger$ \\
    
    \bottomrule
    \end{tabular}
}

\caption{Zero- and 10-shot prompting performance of LLaMA-2 13B model. \fcolorbox{white}{light green}{Green}: best performance per language pair; \fcolorbox{white}{light red}{Red}: worse performance than the original baseline. We test the statistically significance of improvements over the original and $\dagger$ marks results with $p$-value $\leq 0.05$.}
\label{tab:llama-13}
\end{table*}

%% file: tables/llama-chat.tex
\definecolor{light green}{rgb}{0.51, 0.859, 0.529}
\definecolor{light red}{rgb}{0.992, 0.678, 0.678}

\begin{table*}
\centering
\resizebox{\textwidth}{!}{%
    \begin{tabular}{l c c c c c c c}
    \toprule
    \multirow{2}{*}{\textbf{Language}} & \multirow{2}{*}{\textbf{Shots}} & \multicolumn{6}{c}{\textbf{BLEU (↑)} / \textbf{TER (↓)} / \textbf{COMET$_\mathrm{DA}$ (↑)}} \\
    & & Original & Generic & Score & MQM & InstructScore & xCOMET \\ \midrule\midrule

    \textit{LLaMA-2 chat 7B} \\ \midrule
    
    \multirow{2}{*}{\textbf{Zh-En}} & 0 & \multirow{2}{*}{0.47 / 0.75 / 0.70} & 0.43$^\dagger$ / 0.69$^\dagger$ / 0.73$^\dagger$ & 0.45$^\dagger$ / 0.66$^\dagger$ / 0.74$^\dagger$ & 0.40$^\dagger$ / 0.70$^\dagger$ / 0.70 & 0.42$^\dagger$ / 0.68$^\dagger$ / 0.73$^\dagger$ & 0.41$^\dagger$ / 0.67$^\dagger$ / 0.73$^\dagger$ \\
    
    & 10 & & 0.48 / 0.65$^\dagger$ / 0.74$^\dagger$ & 0.48 / 0.64$^\dagger$ / 0.74$^\dagger$ & 0.48 / 0.64$^\dagger$ / 0.73$^\dagger$ & 0.48$^\dagger$ / \textbf{0.63} / 0.75$^\dagger$ & \textbf{0.50} / \textbf{0.63}$^\dagger$ / \textbf{0.76}$^\dagger$ \\ \midrule
    
    \multirow{2}{*}{\textbf{En-De}} & 0 & \multirow{2}{*}{0.45 / 0.86 / 0.69} & 0.42 / 0.68$^\dagger$ / 0.73$^\ddagger$ & 0.51$^\dagger$ / 0.63$^\dagger$ / 0.76$^\dagger$ & 0.54$^\dagger$ / 0.62$^\dagger$ / 0.76$^\dagger$ & - & 0.50$^\dagger$ / 0.6$^\dagger$ / 0.77$^\dagger$ \\
    
    & 10 & & 0.54$^\dagger$ / 0.60$^\dagger$ / 0.78$^\dagger$ & 0.54$^\dagger$ / 0.60$^\dagger$ / 0.78$^\dagger$ & 0.54$^\dagger$ / 0.59$^\dagger$ / 0.78$^\dagger$ & - & \textbf{0.56}$^\dagger$ / \textbf{0.57}$^\dagger$ / \textbf{0.79}$^\dagger$ \\ \midrule
    
    \multirow{2}{*}{\textbf{En-Ru}} & 0 & \multirow{2}{*}{0.43 / 0.82 / 0.73} & 0.40$^\dagger$ / 0.72$^\dagger$ / 0.72$^\ddagger$ & 0.45$^\dagger$ / 0.66$^\dagger$ / 0.73 & 0.48$^\dagger$ / 0.64$^\dagger$ / 0.74$^\ddagger$ & - & 0.45$^\dagger$ / 0.63$^\dagger$ / 0.76$^\dagger$ \\
    
    & 10 & & 0.50$^\dagger$ / 0.61$^\dagger$ / 0.78$^\dagger$ & 0.49$^\dagger$ / 0.62$^\dagger$ / 0.77$^\dagger$ & 0.50$^\dagger$ / 0.61$^\dagger$ / 0.77$^\dagger$ & - & \textbf{0.52}$^\dagger$ / \textbf{0.59}$^\dagger$ / \textbf{0.81}$^\dagger$ \\ \midrule\midrule

    \textit{LLaMA-2 chat 13B} \\ \midrule

    \multirow{2}{*}{\textbf{Zh-En}} & 0 & \multirow{2}{*}{0.47 / 0.75 / 0.70} & 0.40$^\dagger$ / 0.71$^\dagger$ / 0.73$^\dagger$ & 0.45$^\dagger$ / 0.69$^\dagger$ / 0.72$^\dagger$ & 0.47$^\dagger$ / 0.72$^\dagger$ / 0.70 & 0.44 / 0.65$^\dagger$ / 0.75$^\dagger$ & 0.47 / 0.64$^\dagger$ / 0.76$^\dagger$ \\
    
    & 10 & & 0.48 / 0.65 / 0.74$^\dagger$ & 0.48$^\ddagger$ / 0.64 / 0.74$^\dagger$ & 0.49$^\dagger$ / 0.62 / 0.74$^\dagger$ & 0.48$^\dagger$ / 0.64 / 0.75$^\dagger$ & \textbf{0.51} / \textbf{0.62} / \textbf{0.76}$^\dagger$ \\ \midrule
    
    \multirow{2}{*}{\textbf{En-De}} & 0 & \multirow{2}{*}{0.45 / 0.86 / 0.69} & 0.53$^\dagger$ / 0.59$^\dagger$ / 0.78$^\dagger$ & 0.50$^\dagger$ / 0.71$^\dagger$ / 0.74$^\dagger$ & 0.48$^\dagger$ / 0.74$^\dagger$ / 0.72 & - & \textbf{0.57}$^\dagger$ / \textbf{0.54}$^\dagger$ / \textbf{0.80}$^\dagger$ \\
    
    & 10 & & 0.53$^\dagger$ / 0.60$^\dagger$ / 0.78$^\dagger$ & 0.55$^\dagger$ / 0.58$^\dagger$ / 0.70$^\dagger$ & 0.55$^\dagger$ / 0.58$^\dagger$ / 0.79$^\dagger$ & - & 0.55$^\dagger$ / 0.58$^\dagger$ / \textbf{0.80}$^\dagger$ \\ \midrule
    
    \multirow{2}{*}{\textbf{En-Ru}} & 0 & \multirow{2}{*}{0.43 / 0.82 / 0.73} & 0.47$^\dagger$ / 0.61$^\dagger$ / 0.77$^\dagger$ & 0.47$^\dagger$ / 0.72$^\dagger$ / 0.74$^\dagger$ & 0.47$^\dagger$ / 0.72$^\dagger$ / 0.74$^\dagger$ & - & \textbf{0.51}$^\dagger$ / \textbf{0.58}$^\dagger$ / \textbf{0.81}$^\dagger$ \\
    
    & 10 & & 0.47$^\dagger$ / 0.62$^\dagger$ / 0.77$^\dagger$ & 0.49$^\dagger$ / 0.60$^\dagger$ / 0.79$^\dagger$ & 0.49$^\dagger$ / 0.60$^\dagger$ / 0.78$^\dagger$ & - & 0.49$^\dagger$ / 0.62$^\dagger$ / \textbf{0.81}$^\dagger$ \\
    
    \bottomrule
    \end{tabular}
}
\caption{\textit{Top rows}: Prompting performance of LLaMA-2 chat 7B model. \textit{Bottom rows}: LLaMA-2 chat 13B model. \textbf{Bold} denotes best performance for each language pair in 7B and 13B. We test the statistically significance of improvements over the original and $\dagger$ marks results with $p$-value $\leq 0.05$ and $\ddagger$ marks results with $p$-value $\leq 0.1$.}
\label{tab:llama-chat}
\end{table*}
\clearpage

%% file: tables/noerror_result.tex
\begin{table*}
\centering
\resizebox{\textwidth}{!}{%
    \begin{tabular}{l c c c c c c c}
    \toprule
    \multirow{2}{*}{\textbf{Language}} & \multirow{2}{*}{\textbf{Size}} & \multicolumn{4}{c}{\textbf{BLEU (↑)} / \textbf{TER (↓)} / \textbf{COMET$_\mathrm{DA}$ (↑)}} \\
    & & Original & Generic & Score & Fine-grained \\ \midrule

    \multirow{2}{*}{\textbf{Zh-En}} & 7B & \multirow{2}{*}{0.66 / 0.53 / 0.85} & 0.61 / 0.56 / 0.82 & 0.62 / 0.55 / 0.82 & 0.61 / 0.56 / 0.82 \\
    & 13B & & 0.62 / 0.56 / 0.82 & 0.62 / 0.56 / 0.82 & 0.62 / 0.56 / 0.82 \\ \midrule

    \multirow{2}{*}{\textbf{En-De}} & 7B & \multirow{2}{*}{0.65 / 0.56 / 0.88} & 0.57 / 0.61 / 0.84 & 0.52 / 0.65 / 0.81 & 0.58 / 0.61 / 0.84 \\
    & 13B & & 0.64 / 0.56 / 0.88 & 0.65 / 0.56 / 0.87 & 0.64 / 0.56 / 0.87 \\ \midrule

    \multirow{2}{*}{\textbf{En-Ru}} & 7B & \multirow{2}{*}{0.62 / 0.58 / 0.92} & 0.51 / 0.68 / 0.85 & 0.51 / 0.66 / 0.84 & 0.56 / 0.64 / 0.87 \\
    & 13B & & 0.61 / 0.60 / 0.91 & 0.62 / 0.58 / 0.91 & 0.61 / 0.59 / 0.91 \\
    
    \bottomrule
    \end{tabular}
}
\caption{Zero-shot prompting performance for instances with no error in their hypothesis translations. \textbf{Original MT hypothesis}: Translation quality from original MQM dataset. Resulting edits lead to small drop in the metrics but they correct stylistic issues such as translationese.}
\label{tab:noerror_result}
\end{table*}

%% file: tables/fine_grained.tex
\begin{table*}
\centering
\resizebox{\textwidth}{!}{%
    
    

    

\begin{tabular}{l c c c c c c c c c c}
    \toprule
    \multirow{2}{*}{\textbf{Language}} & \multirow{2}{*}{\textbf{Component}} & \multicolumn{3}{c}{\textbf{BLEU (↑)}} & \multicolumn{3}{c}{\textbf{TER (↓)}} & \multicolumn{3}{c}{\textbf{COMET$_\mathrm{DA}$ (↑)}} \\
    \cmidrule(lr){3-5} \cmidrule(lr){6-8} \cmidrule(lr){9-11}
    & & MQM & InstructScore & xCOMET & MQM & InstructScore & xCOMET & MQM & InstructScore & xCOMET \\ \midrule
    
    \multirow{4}{*}{\textbf{Zh-En}} & All & 0.47 & 0.43 & 0.41 & 0.72 & 0.66 & \textbf{0.64} & 0.70 & 0.73 & \textbf{0.72} \\
    & Span & 0.47 & 0.41 & 0.41 & 0.71 & 0.67 & 0.66 & 0.71 & 0.72 & 0.72 \\
    & Type & 0.47 & 0.43 & - & 0.70 & \textbf{0.62} & - & 0.72 & \textbf{0.74} & - \\
    & Severity & \textbf{0.48} & \textbf{0.44} & \textbf{0.44} & \textbf{0.66} & 0.65 & \textbf{0.64} & 0.70 & \textbf{0.74} & \textbf{0.74} \\ \midrule
    
    \multirow{4}{*}{\textbf{En-De}} & All & 0.47 & - & 0.54 & 0.75 & - & 0.60 & 0.71 & - & 0.75 \\
    & Span & 0.49 & - & 0.56 & \textbf{0.71} & - & 0.58 & \textbf{0.72} & - & 0.75 \\
    & Type & 0.49 & - & - & 0.73 & - & - & 0.71 & - & - \\
    & Severity & \textbf{0.50} & - & \textbf{0.56} & \textbf{0.71} & - & \textbf{0.57} & 0.71 & - & \textbf{0.76} \\ \midrule

    \multirow{4}{*}{\textbf{En-Ru}} & All & 0.43 & - & 0.48 & 0.77 & - & 0.62 & 0.74 & - & 0.76 \\
    & Span & \textbf{0.45} & - & 0.48 & 0.75 & - & 0.62 & 0.73 & - & 0.77 \\
    & Type & 0.44 & - & - & 0.76 & - & - & \textbf{0.75} & - & - \\
    & Severity & \textbf{0.45} & - & \textbf{0.50} & 0.76 & - & \textbf{0.61} & \textbf{0.76} & - & \textbf{0.78} \\
    
    \bottomrule
    \end{tabular}
}
\caption{Zero-shot prompting performance of LLaMA-2 7B when breaking down fine-grained feedback into three components. Note that results are missing as InstructScore only supports Zh-En MT pair and xCOMET does not output error type. While the individual contribution of each component is trivial, providing solely the severity level information outperforms the case of giving all components simultaneously. \textbf{Bold} indicates the best performance for each annotation.}
\label{tab:fine_grained}
\end{table*}

%% file: tables/translate_from_scratch.tex
\begin{table*}
\centering
\resizebox{\textwidth}{!}{%
\begin{tabular}{l c c c c}
    \toprule
    \textbf{Language} & \textbf{Type} & \textbf{BLEU (↑)} & \textbf{TER (↓)} & \textbf{COMET$_\mathrm{DA}$ (↑)} \\ \midrule

    \multirow{3}{*}{\textbf{Zh-En}} & Original MQM translation & 0.47 & 0.75 & 0.70 \\
    & LLaMA-2 (7B) & 0.24 & 1.54 & 0.71 \\
    & LLaMA-2 (13B) & 0.47 & 0.72 & 0.73 \\ \midrule

    \multirow{3}{*}{\textbf{En-De}} & Original MQM translation & 0.45 & 0.86 & 0.69 \\
    & LLaMA-2 (7B) & 0.32 & 1.15 & 0.70 \\
    & LLaMA-2 (13B) & 0.50 & 0.68 & 0.73 \\ \midrule

    \multirow{3}{*}{\textbf{En-Ru}} & Original MQM translation & 0.43 & 0.82 & 0.73 \\
    & LLaMA-2 (7B) & 0.23 & 1.25 & 0.66 \\
    & LLaMA-2 (13B) & 0.44 & 0.73 & 0.75 \\
    
    \bottomrule
    \end{tabular}
}
\caption{Zero-shot translation from scratch performance of LLaMA-2 7B and 13B. Results do not support the hypothesis of train/test data contamination.}
\label{tab:translate_from_scratch}
\end{table*}

%% file: tables/qualitative_zhen.tex
\begin{CJK*}{UTF8}{gbsn}

\begin{table*}[!htp]
\centering
\resizebox{\textwidth}{!}{%
    \begin{tabular}{l l}
    \toprule
    \textbf{Type} & \textbf{Example} \\ \midrule
    
    \textbf{Source} & 可以先豁免，到你们调查完再收取我的月费，这样行吗？ \\
    \textbf{Translation} & You can first exempt, and then charge my monthly fee \underline{until} you investigate, okay? \\
    \textbf{Reference} & It can be exempted, and you can charge me with the monthly fee after the investigation, is that Okay? \\ \midrule
    \textbf{Generic} & Can you first exempt me, and then charge my monthly fee until you investigate, okay? \\
    \textbf{Score} & You can first exempt, and then charge my monthly fee until you investigate, okay? \\
    \textbf{MQM} & You can first waive the fee, and then charge my monthly fee after you investigate, okay? \\
    \textbf{InstructScore} & You can first exempt, and then charge my monthly fee until you investigate, okay? \\
    \textbf{xCOMET} & Can you first exempt, and then charge my monthly fee after you investigate, okay? \\
    \textbf{FT} & It can be waived, and you can charge me the monthly fee after you finish the investigation, can that work? \\
    \bottomrule
    \end{tabular}
}
\caption{Example of error in Chinese-English translation pair. \underline{Underlined} part represents the error span in the candidate translation. Each row from Generic to FT (fine-tuning) presents the generations by incorporating each type of feedback. For prompting results, we use 10-shot prompting of LLaMA-2 7B.}
\label{tab:qualitative_zhen}
\end{table*}

\end{CJK*}

%% file: tables/qualitative_ende.tex
\begin{table*}[htbp]
\centering
\resizebox{\textwidth}{!}{%
    \begin{tabular}{l l}
    \toprule
    \textbf{Type} & \textbf{Example} \\ \midrule
    
    \textbf{Source} & Robust Character Creation Tools - Make the character YOU want to play. \\
    \textbf{Translation} & \underline{Robust Character Creation Tools} - Machen Sie den Charakter, den Sie spielen möchten. \\
    \textbf{Reference} & Robuste Charaktererstellungs-Tools - baue dir die Charaktere, die DU spielen möchtest. \\ \midrule
    \textbf{Generic} & Robuste Charaktererstellungswerkzeuge - Machen Sie den Charakter, den Sie spielen möchten. \\
    \textbf{Score} & Robuste Charakter-Erstellungswerkzeuge - Erstellt den Charakter, den Sie spielen möchten. \\
    \textbf{MQM} & Robuste Charakter-Erstellungswerkzeuge - Machen Sie den Charakter, den Sie spielen möchten. \\
    \textbf{xCOMET} & Robuste Charaktererstellung - Machen Sie den Charakter, den Sie spielen möchten. \\
    \textbf{FT} & Robuste Charaktererstellungstools - Machen Sie den Charakter, den Sie spielen möchten. \\
    \bottomrule
    \end{tabular}
}
\caption{Example of major accuracy (mistranslation) error in English-German translation pair. \underline{Underlined} part represents the error span in the candidate translation.}
\label{tab:qualitative_ende}
\end{table*}

%% file: tables/qualitative_enru.tex
\begin{table*}[htbp]
\centering
\resizebox{\textwidth}{!}{%
    \begin{tabular}{l l}
    \toprule
    \textbf{Type} & \textbf{Example} \\ \midrule
    
    \textbf{Source} & Have you check on your doorstep or reception area? \\
    \textbf{Translation} & \russian{Вы проверяли порог своей двери или приемную?} \\
    \textbf{Reference} & \russian{Вы проверили свой порог или \underline{стойку регистрации}?} \\ \midrule
    \textbf{Generic} & \russian{Вы проверили, если вы находитесь на пороге или в регистрационной лавке?} \\
    \textbf{Score} & \russian{Вы проверили свой порог или стойку регистрации?} \\
    \textbf{MQM} & \russian{Вы проверили ваш порог или регистрационную стойку?} \\
    \textbf{xCOMET} & \russian{Вы проверили ли свой порог или стойку регистрации?} \\
    \textbf{FT} & \russian{Проверьте на своем входе или зоне приемных залов?} \\
    \bottomrule
    \end{tabular}
}
\caption{Example of error in English-Russian translation pair. \underline{Underlined} part represents the error span in the candidate translation.}
\label{tab:qualitative_enru}
\end{table*}

%% file: figures/human_intro.tex
\begin{figure*}
    \centering
    \fbox{
        \includegraphics[width=\textwidth]{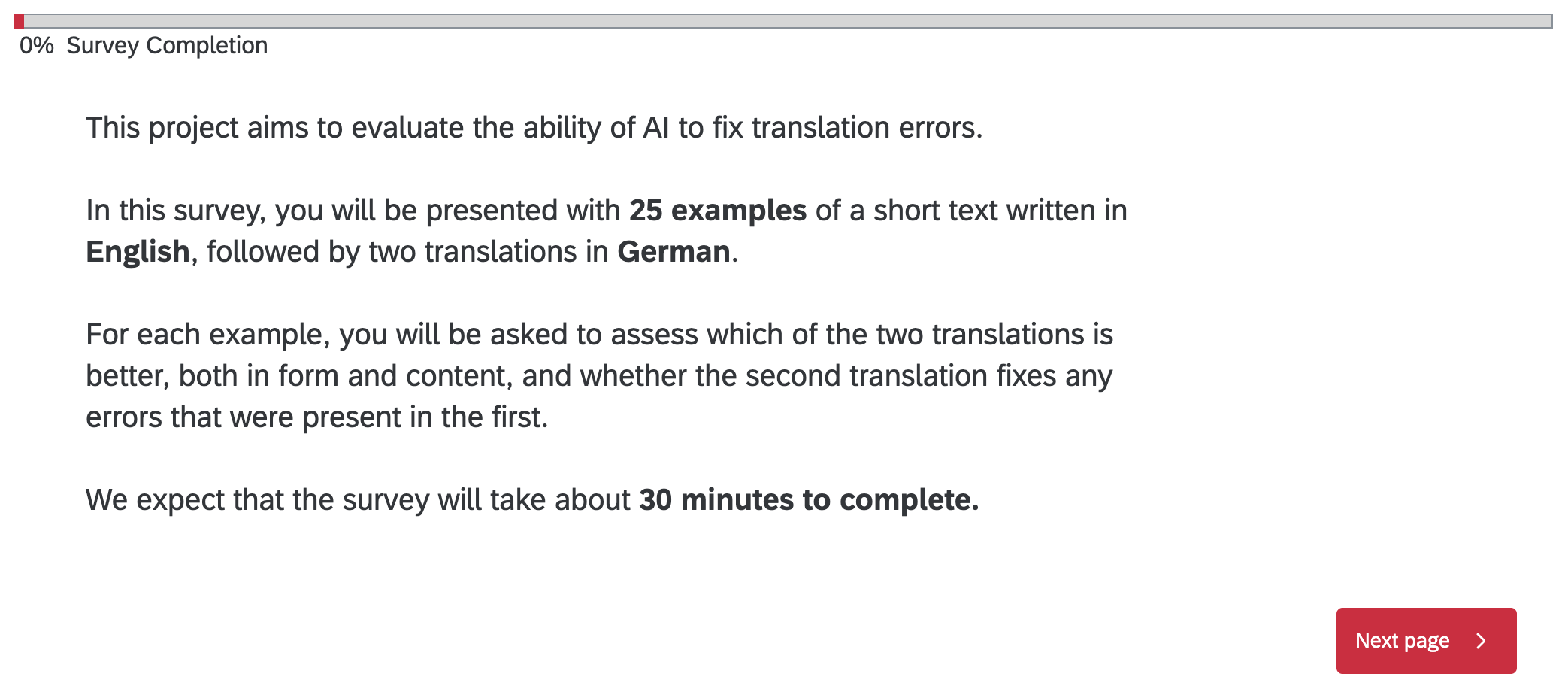}
    }
    \caption{Instructions for human evaluation. This is shown as the first page of our survey to all annotators.} 
    \label{fig:human_intro}
\end{figure*}

%% file: figures/human_example.tex
\begin{figure*}
    \centering
    \fbox{%
        \includegraphics[width=\textwidth]{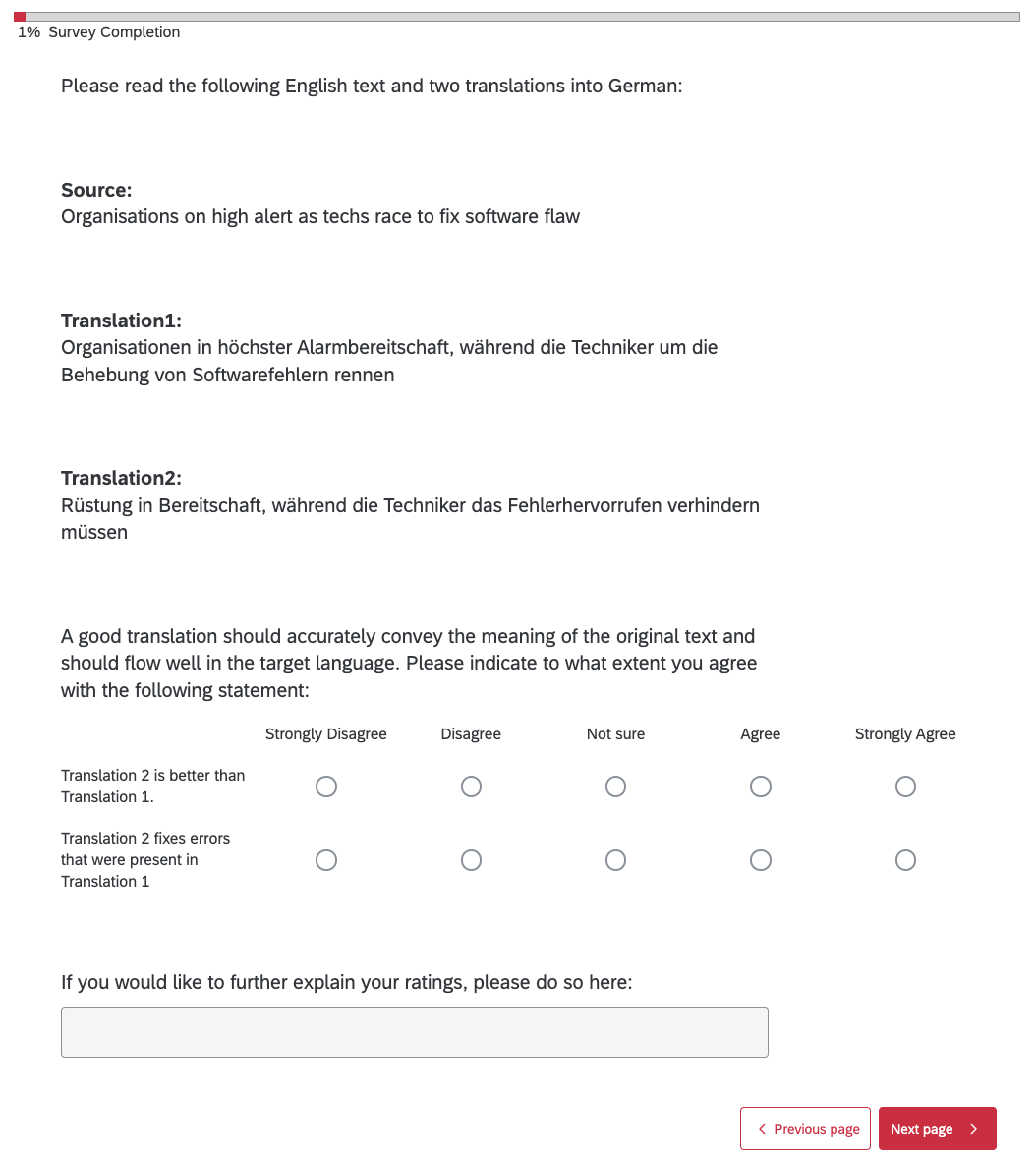}
    }
    \caption{Survey content for human evaluation. Given \textbf{Source}, \textbf{Translation 1} (original translation), and \textbf{Translation 2} (output translation from the bilingual fine-tuned model), annotators are asked to answer 2 questions on a scale from 0 to 5. Extra text box is given for each example for further suggestions.} 
    \label{fig:human_example}
\end{figure*}

%% file: tables/human_eval_comments.tex
\definecolor{light green}{rgb}{0.51, 0.859, 0.529}
\definecolor{light red}{rgb}{0.992, 0.678, 0.678}

\begin{table*}
\centering
\resizebox{\textwidth}{!}{%
    \begin{tabular}{l l}
    \toprule
    \textbf{Language} & \textbf{Feedback} \\ \midrule
    
    \multirow{6}{*}{\textbf{Zh-En}} & (1) Translation 1 is better because it explains more, but Translation 2 corrects the errors that 1 has. \\
    & (2) Translation 1 is better than Translation 2, is more specific and understandable. \\
    & (3) Translation 2 is better because it is easy to understand and \textbf{explains the context}. \\
    & (4) Translation 2 \textbf{fixes all errors} in Translation 1. \\
    & (5) Translation 2 is better because there are \textbf{no errors} and it is more concrete. \\
    & (6) Translation 2 is better, but it can improve more. \\ \midrule
    
    \multirow{2}{*}{\textbf{En-De}} & (1) Although Translation 1 is more faithful to the original source sentence, it looks like it was directly translated from it. \\
    & (2) Translation 2 is \textbf{more fitting to the actual use of the German language syntax and flow.} \\ \midrule

    \multirow{6}{*}{\textbf{En-Ru}} & (1) Translation 2 is better because it \textbf{avoids the major error} of Translation 1. \\
    & (2) Translation 2 is more \textbf{accurate} and \textbf{flows better} in the target language. \\
    & (3) Translation 2 correctly uses the phrase in the source sentence while Translation 1 has a small error, which is not contextually correct. \\
    & (4) Translation 1 does not have a Russian translation of the English text. \\
    & (5) Translation 2 was more emotive than the original text. \\
    & (6) Translation 1 is misleading whereas Translation 2 \textbf{speaks on the actual events.} \\
    
    \bottomrule
    \end{tabular}
}
\caption{Feedback from the human annotators. We refer to \textit{Translation 1} as initial translation and \textit{Translation 2} as output translation from our fine-tuned model.}
\label{tab:human_eval_comments}
\end{table*}
